\pdfoutput=1

\documentclass[11pt]{article}

\usepackage[]{EMNLP2022}

\usepackage{times}
\usepackage{latexsym}

\usepackage[T1]{fontenc}

\usepackage[utf8]{inputenc}

\usepackage{microtype}

\usepackage{inconsolata}

\usepackage{times}
\usepackage{latexsym}
\usepackage{comment}
\usepackage{amsmath,amsfonts,amssymb,amsthm,mathtools}
\usepackage{comment}
\usepackage{graphicx}
\usepackage{xcolor}
\usepackage{color,soul}
\usepackage{booktabs}
\usepackage{mathtools}
\usepackage{linguex}
\usepackage{subfigure}
\usepackage{array}
\usepackage{float}
\usepackage{bm}
\usepackage{arydshln}
\usepackage{tikz}
\usepackage{multirow}

\definecolor{colorpath1}{HTML}{999999}
\definecolor{colorpath2}{HTML}{1764AB}
\definecolor{colorpath3}{HTML}{E27878}
\definecolor{yellowpaper}{HTML}{D6B656}
\definecolor{bluepaper}{HTML}{6C8EBF}
\definecolor{greypaper}{HTML}{5C5C5C}
\DeclareRobustCommand\dashed{\tikz[baseline=-0.6ex]\draw[thick,dashed] (0,0)--(0.35,0);}

\definecolor{alti}{HTML}{1f77b4}
\definecolor{grad}{HTML}{ff7f0e}
\definecolor{ig_l2}{HTML}{2ca02c}
\definecolor{ig_mean}{HTML}{d62728}
\definecolor{grad_input_l2}{HTML}{9467bd}
\definecolor{grad_input_mean}{HTML}{8c564b}
\definecolor{rollout}{HTML}{e377c2}
\definecolor{norm2}{HTML}{7f7f7f}
\definecolor{ours2}{HTML}{bcbd22}

\usepackage[T1]{fontenc}

\usepackage[utf8]{inputenc}

\usepackage{microtype}
\newcommand\norm[1]{\left\lVert#1\right\rVert}

\definecolor{color0}{HTML}{FFFFFF}
\definecolor{color1}{HTML}{D0E1F2}
\definecolor{color2}{HTML}{94C4DF}
\definecolor{color3}{HTML}{4A98C9}
\definecolor{color4}{HTML}{1764AB}

\newcommand*{\mybox}[2]{{\setlength{\fboxsep}{0.15pt}\colorbox{#1}{\strut #2}}}

\usepackage{nameref}
\usepackage{cleveref}

\crefformat{section}{\S#2#1#3}
\crefformat{subsection}{\S#2#1#3}
\crefformat{subsubsection}{\S#2#1#3}
\crefformat{paragraph}{\P#2#1#3}
\crefformat{subparagraph}{\P#2#1#3}
\crefmultiformat{section}{\S#2#1#3}{ and~\S#2#1#3}{, \S#2#1#3}{, and~\S#2#1#3}
\crefmultiformat{subsection}{\S#2#1#3}{ and~\S#2#1#3}{, \S#2#1#3}{, and~\S#2#1#3}
\crefmultiformat{subsubsection}{\S#2#1#3}{ and~\S#2#1#3}{, \S#2#1#3}{, and~\S#2#1#3}
\crefmultiformat{paragraph}{\P\P#2#1#3}{ and~#2#1#3}{, #2#1#3}{, and~#2#1#3}
\crefmultiformat{subparagraph}{\P\P#2#1#3}{ and~#2#1#3}{, #2#1#3}{, and~#2#1#3}
\crefrangeformat{section}{\mbox{\S\S#3#1#4--#5#2#6}}
\crefrangeformat{subsection}{\mbox{\S\S#3#1#4--#5#2#6}}
\crefrangeformat{subsubsection}{\mbox{\S\S#3#1#4--#5#2#6}}
\crefrangeformat{paragraph}{\mbox{\P\P#3#1#4--#5#2#6}}
\crefrangeformat{subparagraph}{\mbox{\P\P#3#1#4--#5#2#6}}
\crefname{part}{Part}{Parts}
\Crefname{part}{Part}{Parts}
\crefname{chapter}{ch.}{ch.}
\Crefname{chapter}{Ch.}{Ch.}
\crefname{footnote}{fn.}{fn.}
\Crefname{footnote}{Fn.}{Fn.}
\crefname{figure}{figure}{figures}
\crefname{subfigure}{figure}{figures}
\Crefname{subfigure}{Figure}{Figures}
\crefname{appsec}{appendix}{appendices}
\Crefname{appsec}{Appendix}{Appendices}
\crefname{algocf}{algorithm}{algorithms}
\Crefname{algocf}{Algorithm}{Algorithms}

\crefname{ExNo}{example}{examples}
\Crefname{ExNo}{Example}{Examples}
\crefname{SubExNo}{example}{examples}
\Crefname{SubExNo}{Example}{Examples}
\crefname{SubSubExNo}{example}{examples}
\Crefname{SubSubExNo}{Example}{Examples}
\crefformat{ExNo}{(#2#1#3)}
\crefformat{SubExNo}{(#2#1#3)}
\crefformat{SubSubExNo}{(#2#1#3)}
\crefrangeformat{ExNo}{\mbox{(#3#1#4--#5#2#6)}}
\crefrangeformat{SubExNo}{\mbox{(#3#1#4--#5#2#6)}}
\crefrangeformat{SubSubExNo}{\mbox{(#3#1#4--#5#2#6)}}
\crefmultiformat{ExNo}{(#2#1#3}{, #2#1#3)}{, #2#1#3}{, #2#1#3)}
\crefmultiformat{SubExNo}{(#2#1#3}{, #2#1#3)}{, #2#1#3}{, #2#1#3)}
\crefmultiformat{SubSubExNo}{(#2#1#3}{, #2#1#3)}{, #2#1#3}{, #2#1#3)}
\crefrangemultiformat{ExNo}{(#3#1#4--#5#2#6}{, #3#1#4--#5#2#6)}{, #3#1#4--#5#2#6}{, #3#1#4--#5#2#6)}
\crefrangemultiformat{SubExNo}{(#3#1#4--#5#2#6}{, #3#1#4--#5#2#6)}{, #3#1#4--#5#2#6}{, #3#1#4--#5#2#6)}
\crefrangemultiformat{SubSubExNo}{(#3#1#4--#5#2#6}{, #3#1#4--#5#2#6)}{, #3#1#4--#5#2#6}{, #3#1#4--#5#2#6)}

\title{Measuring the Mixing of Contextual Information in the Transformer}

\author{Javier Ferrando, Gerard I. Gállego \and Marta R. Costa-jussà \\
         TALP Research Center, Universitat Politècnica de Catalunya \\
         \texttt{\{javier.ferrando.monsonis,gerard.ion.gallego,marta.ruiz\}@upc.edu}}

\begin{document}
\maketitle
\begin{abstract}
The Transformer architecture aggregates input information through the self-attention mechanism, but there is no clear understanding of how this information is mixed across the entire model. Additionally, recent works have demonstrated that attention weights alone are not enough to describe the flow of information. In this paper, we consider the whole attention block --multi-head attention, residual connection, and layer normalization-- and define a metric to measure token-to-token interactions within each layer. Then, we aggregate layer-wise interpretations to provide input attribution scores for model predictions. Experimentally, we show that our method, ALTI (Aggregation of Layer-wise Token-to-token Interactions), provides more faithful explanations and increased robustness than gradient-based methods.
\end{abstract}

\section{Introduction}
The Transformer \cite{NIPS2017_3f5ee243} has become ubiquitous in different tasks across multiple domains, becoming the architecture of choice for many NLP \cite{devlin-etal-2019-bert,NEURIPS2020_1457c0d6} and computer vision \cite{dosovitskiy2021an} tasks.
The self-attention mechanism inside the Transformer is in charge of combining contextual information in its intermediate token representations. Attention weights offer a straightforward layer-wise interpretation, as they provide a distribution over input units, which is often presented as giving the relative importance of each input.

A prominent line of research has investigated the faithfulness of attention weights \cite{jain-wallace-2019-attention,serrano-smith-2019-attention,pruthi-etal-2020-learning, wiegreffe-pinter-2019-attention,madsen2021posthoc} with contradictory conclusions. Some works have studied layer-wise attention patterns by analyzing standard attention \cite{kovaleva-etal-2019-revealing,clark-etal-2019-bert,vig-belinkov-2019-analyzing} and effective attention \cite{Brunner2020On,sun-marasovic-2021-effective}, but explaining the Transformer beyond attention weights needs further investigation \cite{lu2021influence}. 

\begin{table}[t!]
\resizebox{0.49\textwidth}{!}{%
\setlength{\tabcolsep}{0.2pt}
\begin{tabular}{ll}
\multicolumn{1}{l}{$\text{Grad}_{\ell_2}$}\\ 
 \mybox{color0}{\strut{went}} \mybox{color0}{\strut{here}} \mybox{color0}{\strut{just}} \mybox{color0}{\strut{before}} \mybox{color0}{\strut{a}} \mybox{color0}{\strut{movie}} \mybox{color0}{\strut{.}} \mybox{color0}{\strut{the}} \mybox{color0}{\strut{service}} \mybox{color1}{\strut{was}} \mybox{color4}{\strut{fast}} \mybox{color0}{\strut{but}} \mybox{color0}{\strut{that}} \mybox{color0}{\strut{'}} \mybox{color1}{\strut{s}} \mybox{color0}{\strut{it}} \mybox{color0}{\strut{.}}\\ \mybox{color0}{\strut{i}} \mybox{color0}{\strut{ordered}} \mybox{color0}{\strut{the}} \mybox{color0}{\strut{mango}} \mybox{color0}{\strut{and}} \mybox{color0}{\strut{shrimp}} \mybox{color1}{\strut{quesadilla}} \mybox{color1}{\strut{.}} \mybox{color0}{\strut{my}} \mybox{color0}{\strut{friend}}\\ \mybox{color0}{\strut{ordered}} \mybox{color0}{\strut{nachos}} \mybox{color0}{\strut{.}} \mybox{color0}{\strut{the}} \mybox{color0}{\strut{food}} \mybox{color0}{\strut{was}} \mybox{color1}{\strut{not}} \mybox{color0}{\strut{good}} \mybox{color0}{\strut{.}} \mybox{color0}{\strut{i}} \mybox{color0}{\strut{and}} \mybox{color0}{\strut{my}} \mybox{color0}{\strut{friend}} \mybox{color0}{\strut{could}}\\ \mybox{color0}{\strut{not}} \mybox{color0}{\strut{finish}} \mybox{color0}{\strut{our}} \mybox{color0}{\strut{food}} \mybox{color0}{\strut{and}} \mybox{color0}{\strut{we}} \mybox{color0}{\strut{had}} \mybox{color0}{\strut{stomach}} \mybox{color1}{\strut{aches}} \mybox{color0}{\strut{immediately}} \mybox{color0}{\strut{.}}\\
\addlinespace
\multicolumn{1}{l}{IG$_{\ell_2}$}\\ 
 \mybox{color4}{\strut{went}} \mybox{color3}{\strut{here}} \mybox{color1}{\strut{just}} \mybox{color1}{\strut{before}} \mybox{color0}{\strut{a}} \mybox{color0}{\strut{movie}} \mybox{color0}{\strut{.}} \mybox{color0}{\strut{the}} \mybox{color0}{\strut{service}} \mybox{color0}{\strut{was}} \mybox{color1}{\strut{fast}} \mybox{color0}{\strut{but}} \mybox{color0}{\strut{that}} \mybox{color0}{\strut{'}} \mybox{color0}{\strut{s}} \mybox{color0}{\strut{it}} \mybox{color0}{\strut{.}}\\ \mybox{color0}{\strut{i}} \mybox{color0}{\strut{ordered}} \mybox{color0}{\strut{the}} \mybox{color1}{\strut{mango}} \mybox{color0}{\strut{and}} \mybox{color0}{\strut{shrimp}} \mybox{color1}{\strut{quesadilla}} \mybox{color0}{\strut{.}} \mybox{color0}{\strut{my}} \mybox{color0}{\strut{friend}}\\ \mybox{color0}{\strut{ordered}} \mybox{color1}{\strut{nachos}} \mybox{color0}{\strut{.}} \mybox{color0}{\strut{the}} \mybox{color0}{\strut{food}} \mybox{color0}{\strut{was}} \mybox{color0}{\strut{not}} \mybox{color0}{\strut{good}} \mybox{color0}{\strut{.}} \mybox{color0}{\strut{i}} \mybox{color0}{\strut{and}} \mybox{color0}{\strut{my}} \mybox{color0}{\strut{friend}} \mybox{color0}{\strut{could}}\\ \mybox{color0}{\strut{not}} \mybox{color0}{\strut{finish}} \mybox{color1}{\strut{our}} \mybox{color0}{\strut{food}} \mybox{color0}{\strut{and}} \mybox{color0}{\strut{we}} \mybox{color0}{\strut{had}} \mybox{color0}{\strut{stomach}} \mybox{color0}{\strut{aches}} \mybox{color0}{\strut{immediately}} \mybox{color0}{\strut{.}}\\
\addlinespace
\multicolumn{1}{l}{ALTI}\\ 
 \mybox{color0}{\strut{went}} \mybox{color0}{\strut{here}} \mybox{color0}{\strut{just}} \mybox{color0}{\strut{before}} \mybox{color0}{\strut{a}} \mybox{color0}{\strut{movie}} \mybox{color0}{\strut{.}} \mybox{color0}{\strut{the}} \mybox{color0}{\strut{service}} \mybox{color0}{\strut{was}} \mybox{color1}{\strut{fast}} \mybox{color1}{\strut{but}} \mybox{color1}{\strut{that}} \mybox{color0}{\strut{'}} \mybox{color0}{\strut{s}} \mybox{color1}{\strut{it}} \mybox{color0}{\strut{.}}\\ \mybox{color0}{\strut{i}} \mybox{color1}{\strut{ordered}} \mybox{color0}{\strut{the}} \mybox{color1}{\strut{mango}} \mybox{color0}{\strut{and}} \mybox{color0}{\strut{shrimp}} \mybox{color2}{\strut{quesadilla}} \mybox{color0}{\strut{.}} \mybox{color0}{\strut{my}} \mybox{color0}{\strut{friend}}\\ \mybox{color0}{\strut{ordered}} \mybox{color0}{\strut{nachos}} \mybox{color0}{\strut{.}} \mybox{color0}{\strut{the}} \mybox{color1}{\strut{food}} \mybox{color0}{\strut{was}} \mybox{color2}{\strut{not}} \mybox{color2}{\strut{good}} \mybox{color0}{\strut{.}} \mybox{color0}{\strut{i}} \mybox{color0}{\strut{and}} \mybox{color0}{\strut{my}} \mybox{color0}{\strut{friend}} \mybox{color0}{\strut{could}}\\ \mybox{color0}{\strut{not}} \mybox{color1}{\strut{finish}} \mybox{color0}{\strut{our}} \mybox{color0}{\strut{food}} \mybox{color0}{\strut{and}} \mybox{color0}{\strut{we}} \mybox{color1}{\strut{had}} \mybox{color3}{\strut{stomach}} \mybox{color4}{\strut{aches}} \mybox{color1}{\strut{immediately}} \mybox{color0}{\strut{.}}\\
\bottomrule
\end{tabular}
}
\caption{Saliency maps of BERT generated by two common gradient methods and by our proposed method, ALTI, for a \textbf{negative} sentiment prediction example of Yelp dataset.} 
\label{tab:qualitative_examples_sst2_bert}
\end{table}

\citet{kobayashi-etal-2020-attention} extended the explainability of the model by also considering the magnitude of the vectors involved in the attention mechanism, and \citet{kobayashi-etal-2021-incorporating} went as far as incorporating the layer normalization and the skip connection in their analysis. While these works have helped better understand the layer-wise behavior of the Transformer, there is a mismatch between layer-wise attention distributions and global input attributions \cite{pascual-etal-2021-telling} since intermediate layers only attend to a mix of input tokens. \citet{Brunner2020On} quantified the aggregation of contextual information throughout the model with a gradient attribution method. Although they found the self-attention mechanism greatly mixes the information of the model input, they were able to recover the token identity from hidden layers with high accuracy with a learned linear mapping. This phenomenon is partially explained by \citet{kobayashi-etal-2021-incorporating} and \citet{lu2021influence}, who have shown the relatively small impact of the multi-head attention, which loses influence with respect to the residual connection, consequently revealing a reduced entanglement of contextual information in BERT. Finally, \citet{abnar-zuidema-2020-quantifying} proposed the attention rollout method, which measures the mixing of information by linearly combining attention matrices, a method that has been extended to Transformers in the visual domain \cite{Chefer_2021_ICCV,Chefer_2021_CVPR}. A drawback of this method is that it assumes an equal influence of the skip connection and the attention weights.

 In this work, we propose ALTI, an interpretability method that provides input tokens relevancies\footnote{We use `relevancies', `attributions', and `importances' interchangeably.} to the model predictions by measuring the aggregation of contextual information across layers. We use the attention block decomposition proposed by \citet{kobayashi-etal-2021-incorporating} and refine the measure of the contribution of each input token representation to the attention block output (layer-wise token-to-token interactions), based on the properties of the representation space and the limitations of previously proposed metrics. We then aggregate the layer-wise explanations and track the mixing of information in each token representation, yielding input attributions for the model predictions. Finally, in the Text Classification and Subject-Verb Agreement tasks, we show ALTI scores higher than gradient-based methods and previous similar approaches in two common faithfulness metrics, while showing greater robustness. The code to reproduce the experiments is publicly available.\footnote{\url{https://github.com/mt-upc/transformer-contributions}.}
 
 \begin{figure}[!t]
\begin{center}
\includegraphics[width=0.37\textwidth]{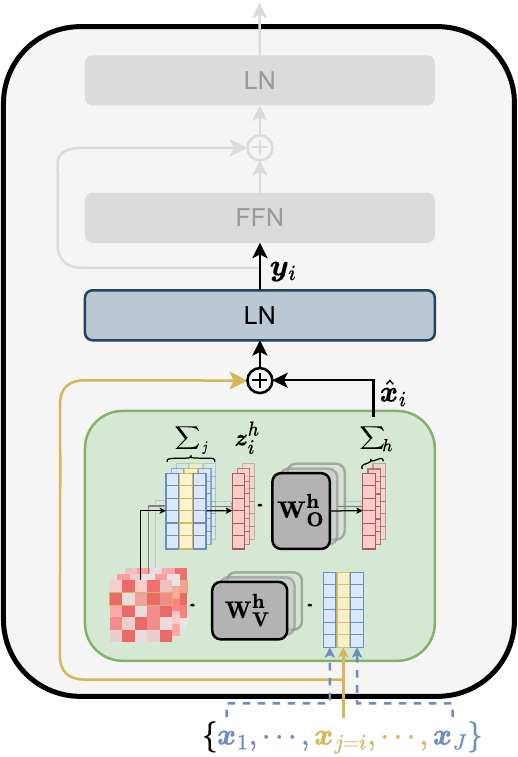}
\caption{Transformer layer with the modules considered in the analysis. We compute layer-wise token-to-token interactions by measuring the contributions of each input token representation $\bm{x}_{j}$ to the attention block output $\bm{y_i}$.}
\label{fig:postnorm_transformer}
\end{center}
\end{figure}

\section{Background}
\subsection{Attention Block Decomposition}
The attention block\footnote{We refer to `attention block' as the multi-head attention, residual connection and layer normalization components.} computations in each layer (highlighted parts in \Cref{fig:postnorm_transformer}) can be reformulated \cite{kobayashi-etal-2021-incorporating} as a simple expression of the layer input representations. Given a sequence of token representations $\mathbf{X} = (\bm{x}_1,\cdots,\bm{x}_{J}) \in \mathbb{R}^{d \times J}$, and a model with $H$ heads and head dimension $d_h = d/H$, the attention block output of the $i$-th token $\bm{y}_i$ is computed by applying the layer normalization (LN) over the sum of the residual vector $\bm{x}_i$, and the output of the multi-head attention module (MHA) $\hat{\bm{x}}_i$:\begin{equation}\label{eq:post_layer_output}
\bm{y}_i = \text{LN}(\hat{\bm{x}}_i + \bm{x}_{i})
\end{equation}
Each head inside MHA computes\footnote{The bias vector associated with $\mathbf{W}_V^{h}$ is omitted for the sake of simplicity.} $\bm{z}^{h}_i \in \mathbb{R}^{d_h}$:
\begin{equation}\label{eq:head_output}
\bm{z}^{h}_i = \sum_j^J  \mathbf{A}_{i,j}^{h} \mathbf{W}_V^{h}\bm{x}_{j}
\end{equation}
with $\mathbf{A}_{i,j}^{h}$ referring to the attention weight where token $i$ attends token $j$, and $\mathbf{W}_V^{h} \in \mathbb{R}^{d_h \times d}$ to a learned weight matrix. $\hat{\bm{x}}_i$ is calculated by concatenating each $\bm{z}^{h}_i$ and projecting the joint vector through $\mathbf{W}_O \in \mathbb{R}^{d \times d}$:
\begin{equation}
\hat{\bm{x}}_i = \mathbf{W}_O\;\text{Concat}(\bm{z}^{1}_i, \cdots, \bm{z}^{H}_i)
\end{equation}
This is equivalent to a sum over heads where each $\bm{z}^{h}_i$ is projected through the partitioned weight matrix $\mathbf{W}_O^{h} \in \mathbb{R}^{d \times d_h}$ and adding the bias $\bm{b}_{O} \in \mathbb{R}^d$:
\begin{equation}\label{eq:mha_output1}
\hat{\bm{x}}_i = \sum^H_h \mathbf{W}_O^{h} \bm{z}^{h}_i + \bm{b}_O
\end{equation}
By swapping summations we can now rewrite Eq.~\ref{eq:post_layer_output} as:
\begin{equation}\label{eq:mha_output2}
\resizebox{0.48\textwidth}{!}{$\displaystyle{
\bm{y}_i = \text{LN}\left(\sum_j^J \sum^H_h \mathbf{W}_O^{h} \mathbf{A}_{i,j}^{h} \mathbf{W}_V^{h}\bm{x}_{j} + \bm{b}_O + \bm{x}_{i}\right)}$}
\end{equation}
\begin{figure}[!t]
\begin{center}\includegraphics[width=0.25\textwidth]{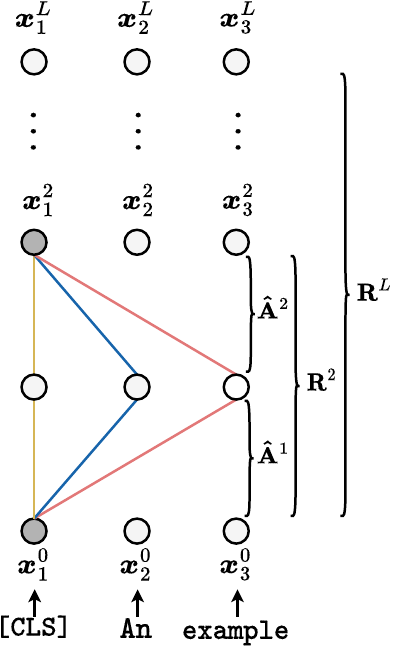}
\caption{Example of attention graph. The relevance $\left(\mathbf{R}^2_{\texttt{[CLS]}}\right)$ of the input token \texttt{[CLS]} token to its second layer representation $\bm{x}^{2}_{1}$ is obtained by summing all possible paths (coloured).}
\label{fig:attention_graph}
\end{center}
\end{figure}
Given a vector $\bm{u}$, $\text{LN}(\bm{u})$ can be reformulated as $\frac{1}{\sigma(\bm{u})}\mathbf{L} \bm{u} + \mathbf{\beta}$ (see \Cref{apx:ln_derivation}), where $\mathbf{L}$ is a linear transformation. Thanks to the linearity of $\mathbf{L}$, we can express $\bm{y}_i$ as:
\begin{equation}\label{eq:post_layer_transformed_vectors}
\bm{y}_i = \sum_j^JT_i(\bm{x}_j) + \frac{1}{\sigma \left(\bm{\hat{x}}_i + \bm{x}_{i}\right)} \mathbf{L}\bm{b}_O + \beta
\end{equation}
where the transformed vectors $T_i(\bm{x}_j)$ are:
\begin{equation*}
\resizebox{0.48\textwidth}{!}{$\displaystyle{
  T_i(\bm{x}_j)=\left\{
  \begin{array}{@{}ll@{}}
    \frac{1}{\sigma \left(\bm{\hat{x}}_i + \bm{x}_{i}\right)}  \mathbf{L} \sum^H_h \mathbf{W}_O^{h} \mathbf{A}_{i,j}^{h} \mathbf{W}_V^{h}\bm{x}_{j} & \mbox{if}~ j \neq i \\
    \frac{1}{\sigma \left(\bm{\hat{x}}_i + \bm{x}_{i}\right)}  \mathbf{L} \left(\sum^H_h \mathbf{W}_O^{h} \mathbf{A}_{i,j}^{h} \mathbf{W}_V^{h}\bm{x}_{j} + \bm{x}_i \right) & \mbox{if}~ j=i
  \end{array}\right.}$}
\end{equation*}
\citet{kobayashi-etal-2021-incorporating} stated that the contribution $c_{i,j}$ of each input vector $\bm{x}_j$ to the layer output $\bm{y}_i$ can be estimated by how much its transformed vector $T_i(\bm{x}_j)$ affects the result in Eq.~\ref{eq:post_layer_transformed_vectors}. They propose using the Euclidean norm of the transformed vector as the metric of contribution:
\begin{equation}\label{eq:kobayashi_contributions}
c_{i,j} = \norm{T_i(\bm{x}_j)}_2
\end{equation}

\subsection{Attention Rollout} \label{sec:rollout}
\citet{abnar-zuidema-2020-quantifying} proposed to measure the mixing of contextual information across the model by relying on attention weights, creating an "attention graph" where nodes represent tokens and hidden representations, and edges attention weights. Two nodes in different layers are connected through multiple paths. To add the residual connection, the attention weights matrix gets augmented with the identity matrix $\mathbf{\hat{A}}^l = 0.5\mathbf{A}^l + 0.5\mathbf{I}$.

We can compute the amount of information flowing from one node to another in different layers by multiplying the edges in each path, and summing over the different paths. In the example of \Cref{fig:attention_graph}, the amount of input information of \texttt{[CLS]} in its second layer representation $\bm{x}^2_{1}$ can be obtained as \textcolor{yellowpaper}{$\mathbf{\hat{A}}^2_{1,1}\cdot\mathbf{\hat{A}}^1_{1,1}$} $+$ \textcolor{colorpath2}{$\mathbf{\hat{A}}^2_{1,2}\cdot\mathbf{\hat{A}}^1_{2,1}$} $+$ \textcolor{colorpath3}{$\mathbf{\hat{A}}^2_{1,3}\cdot\mathbf{\hat{A}}^1_{3,1}$}. This is equivalent to the dot product between $\mathbf{\hat{A}}^2_{1,:}$ and $\mathbf{\hat{A}}^1_{:,1}$, which generalizes to the matrix multiplication when considering all tokens, giving the input relevance matrix at layer $l$, $\mathbf{R}^l \in \mathbb{R}^{J \times J}$:
\begin{equation}\label{eq:attention_rollout}
\mathbf{R}^l = \mathbf{\hat{A}}^{l} \cdot \mathbf{\hat{A}}^{l-1} \cdot \; \cdots \; \cdot \mathbf{\hat{A}}^{1}
\end{equation}
\section{Proposed Approach}
\begin{figure}[!t]
    \begin{centering}
    \includegraphics[width=0.48\textwidth]{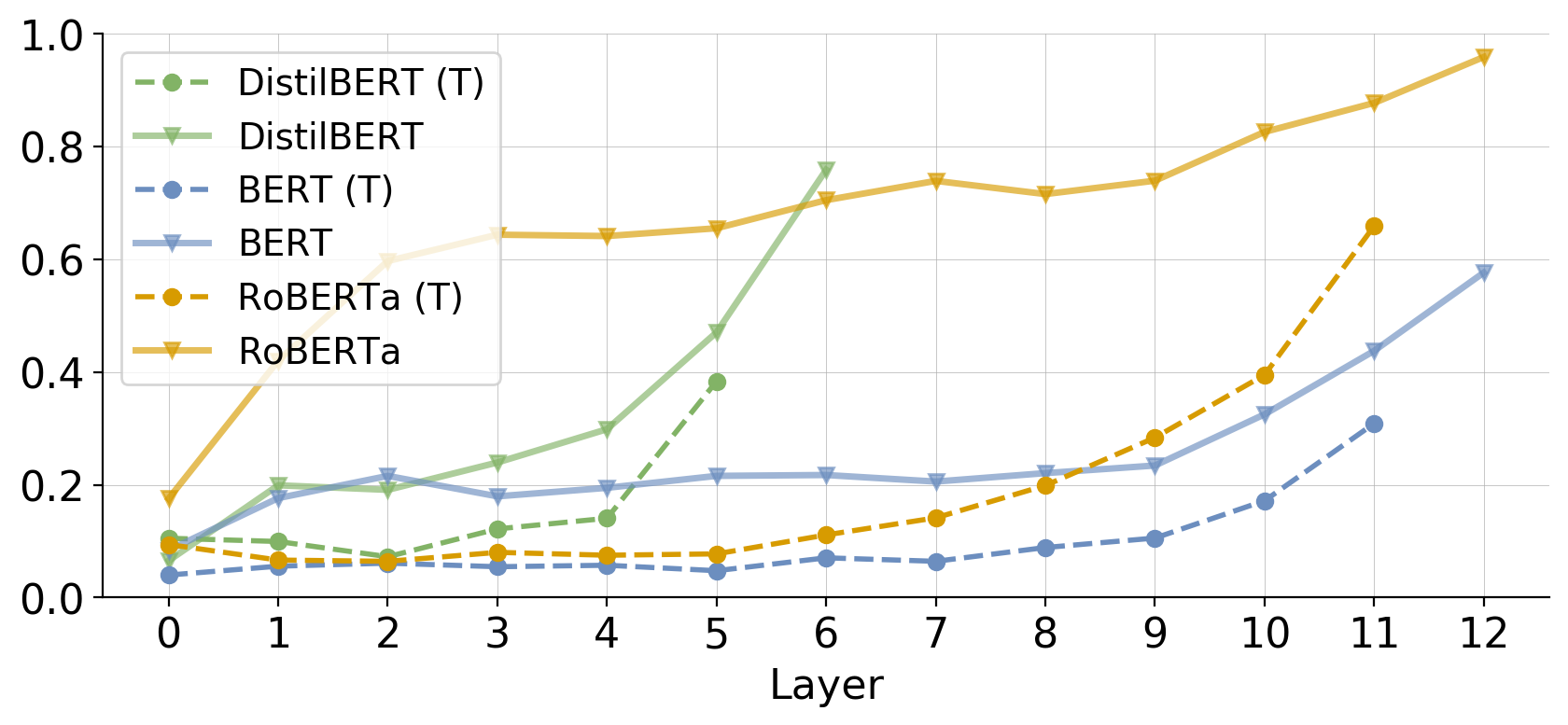}
    \caption{Average cosine similarity of attention block output representations (solid line) and transformed representations (dashed line) in 500 random samples of SST-2 dataset.}
    \label{fig:mean_cosine_similarity}
    \end{centering}
\end{figure}

\begin{figure*}[t]
\begin{center}
\includegraphics[width=0.75\textwidth]{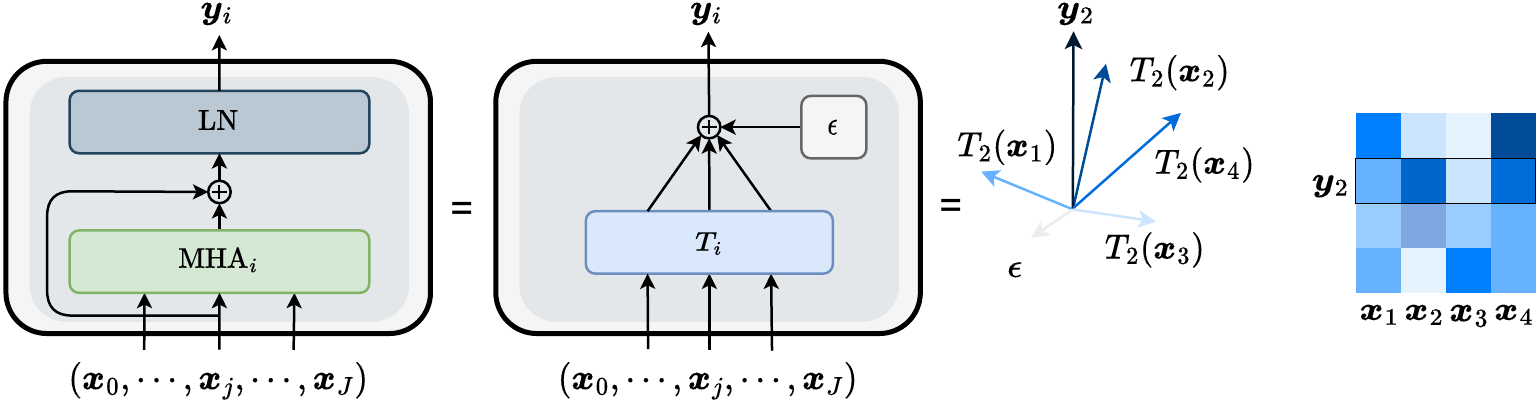}
\caption{The self-attention block (left) at each position $i$ can be decomposed as a summation of transformed input vectors (middle). The closest vector ($T_{2}(\bm{x}_2)$) \textit{contributes} the most to $\bm{y}_{2}$. We obtain a matrix of contributions $\mathbf{C}$ (right) reflecting layer-wise token-token interactions.}
\label{fig:scheme}
\end{center}
\end{figure*}

The decomposition of the attention block, represented as a sum of vectors in Eq.~\ref{eq:post_layer_transformed_vectors}, allows us to interpret token-to-token interactions within each layer. \citet{kobayashi-etal-2021-incorporating} proposed to measure the influence of each input token with the $\ell_2$ norm of the transformed vectors (Eq.~\ref{eq:kobayashi_contributions}). We present two reasons why this estimation may not be accurate:

\begin{enumerate}

\item A property of the contextual representations in Transformer-based models is that they are highly anisotropic \cite{ethayarajh-2019-contextual}, i.e. the expected cosine similarity of randomly sampled token representations tend to be close to 1 (solid lines in \Cref{fig:mean_cosine_similarity}). However, transformed representations exhibit reduced anisotropy, especially for the first layers, where there is almost isotropy (dashed lines in \Cref{fig:mean_cosine_similarity}), i.e. they are more randomly spread across the space. This reinforces the need of accounting for the vector's orientation in space, as opposed to solely relying on their norm.\label{item:low_anisotropy}

\item Recent studies \cite{kovaleva-etal-2021-bert,DBLP:conf/iclr/CaiHB021,luo-etal-2021-positional} have found that some embedding dimensions acquire disproportionately large values, dominating the similarity measures \cite{timkey-van-schijndel-2021-bark}. $\ell_2$ normalized metrics, since they square each vector component, unavoidably weigh heavily the outlier dimensions.\label{item:large_dimensions}
\end{enumerate}

We can analyze the expression in Eq.~\ref{eq:post_layer_transformed_vectors} as $T_i(\bm{x}_j)$ vectors \textit{contributing} to the sum resultant $\bm{y}_i$. We propose to measure how much each transformed vector contributes to the sum by means of its distance to the output vector $\bm{y}_i$. We expect that the closer the vector is to $\bm{y}_i$, the higher its contribution (\Cref{fig:scheme}). In this way, we take into account where each transformed vector lies in the representation space (Reason \ref{item:low_anisotropy}). Due to its robustness to the aforementioned idiosyncratic dimensions (Reason \ref{item:large_dimensions}), we use $\ell_1$ norm, i.e. the Manhattan distance between the attention block output and the transformed vector:
\begin{equation}\label{eq:distance}
d_{i,j} = \norm{\bm{y}_i - T_i(\bm{x}_j)}_1
\end{equation}
The level of contribution of $\bm{x}_j$ to $\bm{y}_i$, $c_{i,j}$, is proportional to the proximity of $T_i(\bm{x}_j)$ to $\bm{y}_i$. The closer the transformed vector is to $\bm{y}_i$, the larger its contribution. We measure proximity as the negative of Manhattan distance $-d_{i,j}$. Finally, we neglect the contributions of those vectors lying beyond the $\ell_1$ length of $\bm{y}_i$: 
\begin{equation}\label{eq:contributions}
c_{i,j} = \frac{\text{max}(0,-d_{i,j} + \norm{\bm{y}_i}_1)}{\sum_k \text{max}(0, -d_{i,k} + \norm{\bm{y}_i}_1)}
\end{equation}
Computing Eq.~\ref{eq:distance} and Eq.~\ref{eq:contributions} for all $\bm{y}_i$ gives us the contributions matrix $\mathbf{C} \in \mathbb{R}^{J \times J}$ containing every token-to-token interaction within the layer.

\begin{figure}[!t]
  \centering
  \begin{tabular}{@{}c@{}}
    \includegraphics[width=.38\textwidth]{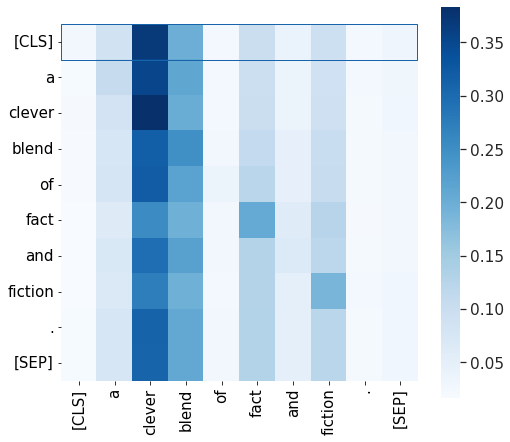} 
  \end{tabular}

  \vspace{\floatsep}

  \begin{tabular}{@{}c@{}}
  \small
 \mybox{color0}{\strut{[CLS]}} \mybox{color0}{\strut{a}} \mybox{color4}{\strut{clever}} \mybox{color2}{\strut{blend}} \mybox{color0}{\strut{of}} \mybox{color1}{\strut{fact}} \mybox{color0}{\strut{and}} \mybox{color1}{\strut{fiction}} \mybox{color0}{\strut{.}} \mybox{color0}{\strut{[SEP]}}\\
  \end{tabular}

  \caption{Top: input (columns) attribution scores $\mathbf{R}^{L}$ provided by ALTI in BERT's last layer token representations (rows). Bottom: attribution scores corresponding to \texttt{[CLS]} token ($\mathbf{R}^L_{\texttt{[CLS]}}$).}
  \label{fig:relevances_bert}
\end{figure}

We also propose to consider contributions across the model as a "contribution graph", similar to the attention graph in Section \Cref{sec:rollout} but using the obtained contributions instead of attention weights. We can then track the amount of contextual information from the input tokens in intermediate token representations, which we use as input attribution scores. By combining linearly the contribution matrices up to layer $l$ (\Cref{fig:relevances_bert} bottom) we get:
\begin{equation}
\mathbf{R}^l = \mathbf{C}^{l} \cdot \mathbf{C}^{l-1} \cdot \; \cdots \; \cdot \mathbf{C}^{1}
\end{equation}
\section{Experimental Setup}
We perform our experiments in the Text Classification (TC) and the Subject-Verb Agreement (SVA) tasks. The former evaluates how models classify an entire input sequence, the latter assesses the ability of a model to capture syntactic phenomena \cite{linzen2016assessing,DBLP:journals/corr/abs-1901-05287}. For the TC task, we use the Stanford Sentiment Treebank v2 (SST-2) \cite{socher-etal-2013-recursive} composed of short sentences, IMDB \cite{maas-etal-2011-learning} with movies reviews and longer inputs than SST2, and Yelp Dataset Challenge \cite{NIPS2015_250cf8b5} containing user's reviews from businesses of similar length than in IMDB. All of them have positive and negative sentiment sentences. For the SVA task, we use \citet{linzen2016assessing} dataset, which includes sentences from Wikipedia containing a present-tense verb that agrees in grammatical number (singular/plural) with the head of the subject. The sentence is fed into the model with its verb masked, and the model is asked to predict if the masked verb is singular or plural\footnote{As a general rule, a singular verb has an `s' added to it in the present tense, such as eats, plays, is, has. A plural verb does not have an `s' added to it.} (binary classification):
\begin{align*}
&\text{\textit{At least four \textbf{players} from the 1983 draft now}}\\
&\underbracket[0.5pt][7pt]{{\texttt{[MASK]}}}_{\text{plural}}\text{\textit{as coaches.}}
\end{align*}
\subsection{Models}
For our experiments we consider three common Transformer pre-trained models\footnote{We use the models available at \url{https://github.com/huggingface/transformers} \cite{wolf-etal-2020-transformers}.} with different sizes and pre-training procedures: BERT \cite{devlin-etal-2019-bert}, DistilBERT \cite{Sanh2019DistilBERTAD} and RoBERTa \cite{DBLP:journals/corr/abs-1907-11692}.

In the TC task, we use fine-tuned models provided by TextAttack \cite{morris2020textattack}. For the robustness analysis in Section \Cref{sec:robustness_analysis}, we fine-tune 10 pre-trained BERT models on SST-2 with the recommended hyperparameters in \citet{devlin-etal-2019-bert}. We compute attribution scores from the row $\mathbf{R}^L_{\texttt{[CLS]}} \in \mathbb{R}^J$ (\Cref{fig:relevances_bert} bottom) that corresponds to the final layer \texttt{[CLS]} embedding, considered a sentence representation for classification tasks. Regarding the SVA task, we split \citet{linzen2016assessing} dataset into 60\%/20\%/20\% for training, validation, and testing respectively, and fine-tune a pre-trained BERT model. We use the input relevances of $\mathbf{R}^{L}_{\texttt{[MASK]}}$.

\subsection{Faithfulness Metrics}\label{sec:faithfulness_metrics}
An interpretation is considered to be faithful if it accurately reflects a model’s decision-making process. A well-established method for measuring faithfulness is by deleting parts of the input sentence $\mathbf{x}$ and observing the change in the predicted probability. Two common erasure-based metrics are comprehensiveness (comp.) and sufficiency (suff.) \citep{deyoung-etal-2020-eraser}. \citet{chan-etal-2022-comparative} have demonstrated that they have higher diagnosticity, i.e. they favor faithful interpretations over randomly generated ones, and lower time complexity than other well-known faithfulness metrics. Comprehensiveness and sufficiency are defined as:

\paragraph{Comprehensiveness.} Measures the change in probability of the predicted class after removing important tokens:
\begin{equation}
{\text{Comp.}} = \frac{1}{|B|+1} \sum_{k \in B} (f(\mathbf{x}) -  f(\mathbf{x} \backslash \mathbf{r}_{:k\%}))
\label{eq:comprehensiveness}
\end{equation}

where $\mathbf{r}_{:k\%}$ refers to the top-$k\%$ most important tokens obtained by an interpretability method. The higher the drop in the probability, the more faithful the interpretation.
\paragraph{Sufficiency.} Captures if important tokens are enough to retain the original prediction:
\begin{equation}
{\text{Suff.}} = \frac{1}{|B|+1} \sum_{k \in B} (f(\mathbf{x}) -  f(\mathbf{r}_{:k\%}))
\label{eq:sufficiency}
\end{equation}  
Lower values of sufficiency indicate a more faithful interpretation, since, in that case, the prediction doesn't change when considering only the important tokens. As in the original paper, for both metrics we use $B = \{0,5,10,20,50\}$.

\subsection{Input Attribution Methods}\label{apx:grad_methods_definition}
Input attribution methods rank input tokens in accordance with how they impact model predictions. They can be divided into: gradient-based methods, perturbation-based, and those relying on the attention mechanism. The gradient of the model's output with respect to the input embeddings is often used as a baseline of faithfulness interpretation \cite{jain-wallace-2019-attention}. \citet{atanasova-etal-2020-diagnostic, zaman2022multilingual} show that gradient-based methods perform better than other interpretability methods, regarding different faithfulness metrics. Finally, perturbation-based methods \cite{Zeiler_occlusion,lime} compute attributions by replacing the original sentence with a modification. \citet{zaman2022multilingual} show that erasure-based methods, such as comprehensiveness and sufficiency favor perturbation-based methods attributed to noise due to the OOD perturbations.

\paragraph{Gradient.}  Considering the model $f$ taking as input a sequence of embeddings $\mathbf{X}^0 \in \mathbb{R}^{d \times J}$, $f$ can be approximated by the linear part of the Taylor-expansion at a baseline point \cite{simonyan_grads}, $f(\mathbf{X}^{0}) \approx \nabla f(\mathbf{X}^{0}) \cdot \mathbf{X}^{0}$. Then, $\nabla f(\mathbf{X}^{0})$ gives a score per embedding dimension, which is often considered as how sensitive the model is to each input dimension when predicting a certain class. To get per token saliency scores \cite{li-etal-2016-visualizing}, we obtain the gradient vector corresponding to the $j$-th token $\nabla_{\bm{x}^{0}_{j}} f(\mathbf{X}^0)= \frac{\partial f(\mathbf{X}^0)}{\partial \bm{x}^0_j}$. Then, we aggregate the gradient vector into a scalar using the $\ell_2$ norm ($\text{Grad}_{\ell_2}$):
\begin{align}\label{eq:l2_norm_grad}
\text{attr}(x_j) &= \norm{\nabla_{\bm{x}_{j}^{0}}f(\mathbf{X}^{0})}_2
\end{align}
Recently, \citet{bastings2021will} showcased (in BERT and SST-2) the high degree of faithfulness of $\text{Grad}_{\ell_2}$ method.
\paragraph{Gradient $\times$ input.}
This method \cite{grad_input2} performs the multiplication of the gradient and the corresponding input embedding. Each component of the gradient vector gets multiplied by the corresponding component of the embedding. Following \cite{atanasova-etal-2020-diagnostic, zaman2022multilingual}, we aggregate the component scores into a single scalar by taking the $\ell_2$ norm ($\text{G}\times\text{I}_{\ell_2}$) as in Eq.~\ref{eq:l2_norm_grad} or by taking the mean ($\text{G}\times\text{I}_{\mu}$) as follows:
\begin{equation}
\text{attr}(x_j) = \frac{1}{N} \sum_{k=1}^{d} |\nabla_{x^{0}_{jk}} f(\mathbf{X}^{0}) \cdot x^{0}_{jk}|
\end{equation}

\paragraph{Integrated Gradients.}
Integrated gradients \cite{pmlr-v70-sundararajan17a} approximates the integral of gradients of the model’s output with respect to the inputs along the straight line path from a baseline input $\bm{B}$, to the actual input. The attribution score for each embedding dimension is defined as:
\begin{equation}\label{eq:ig}
(x^{0}_{jk}-\bm{b}_{jk}) \cdot \frac{1}{m} \sum_{c=1}^{m}\nabla_{\hat{x}^{0}_{jk}}f(\hat{\bm{X}}^{0}_{c})
\end{equation}
where $\hat{\bm{X}}^{0}_{c}=\bm{B}+\frac{c}{m}(\bm{X}^{0}-\bm{B})$, and $m$ number of steps. As baseline, we use repeated \texttt{[MASK]} vectors for each word except for $\texttt{[CLS]}$ and $\texttt{[SEP]}$ \cite{sajjad-etal-2021-fine}, and 100 steps. We aggregate ($\text{IG}_{\ell_2}$ and $\text{IG}_{\mu}$) the attribution scores of the embedding dimensions of Eq.~\ref{eq:ig} to obtain $\text{attr}(x_j)$.

Finally, we normalize the obtained attribution scores in the range so that they sum 1. We use the Captum library implementations \cite{kokhlikyan2020captum}.

\paragraph{Attention.} Attention-based methods that provide input attributions include Attention \textit{Rollout} \cite{abnar-zuidema-2020-quantifying}, as described in Section \Cref{sec:rollout}. Concurrent to this work, \citet{globenc} propose \textit{Globenc}, which combines Attention Rollout aggregation technique with \citet{kobayashi-etal-2021-incorporating} layer-wise contributions (Eq.~\ref{eq:kobayashi_contributions}), with the addition of the layer normalization of the FFN module. In Section \Cref{sec:ln2_addition} we compare ALTI to Globenc.

\begin{table*}[t]
\renewcommand{\arraystretch}{0.9}
\centering
\small
\resizebox{\textwidth}{!}{%
\begin{tabular}{m{2.25cm}m{.75cm}<{\centering}m{.75cm}<{\centering}m{.75cm}<{\centering}m{.85cm}<{\centering}m{.75cm}<{\centering}m{.75cm}<{\centering}m{.75cm}<{\centering}m{.85cm}<{\centering}m{.75cm}<{\centering}m{.75cm}<{\centering}m{.75cm}<{\centering}m{.85cm}<{\centering}}
  \toprule
  & \multicolumn{8}{c}{\bf{BERT}}                                                            & \multicolumn{2}{c}{\bf{RoBERTa}} & \multicolumn{2}{c}{\bf{DistilBERT}} \\
  \cmidrule{2-3}\cmidrule{4-5}\cmidrule{6-7}\cmidrule{8-9}\cmidrule{10-11}\cmidrule{12-13}
  & \multicolumn{2}{c}{\bf{SST-2}} & \multicolumn{2}{c}{\bf{Yelp}} & \multicolumn{2}{c}{\bf{IMDB}} & \multicolumn{2}{c}{\bf{SVA}} & \multicolumn{2}{c}{\bf{SST-2}} & \multicolumn{2}{c}{\bf{SST-2}}
\\
\cmidrule{2-3}\cmidrule{4-5}\cmidrule{6-7}\cmidrule{8-9}\cmidrule{10-11}\cmidrule{12-13}
  \bf{Methods} & \bf{Comp.$\uparrow$}  & \bf{Suff.$\downarrow$} & \bf{Comp.}  & \bf{Suff.} & \bf{Comp.}  & \bf{Suff.} & \bf{Comp.}  & \bf{Suff.} & \bf{Comp.}  & \bf{Suff.} & \bf{Comp.}  & \bf{Suff.}\\
  \midrule
  $\text{Grad}_{\ell_2}$ \textcolor{grad}{\dashed}& 0.204   & 0.076    & 0.083    & 0.101    & 0.192    & 0.052    & 0.284   & 0.145 & 0.190 &0.075 & 0.230 & 0.066 \cr
  $\text{IG}_{\ell_2}$ \textcolor{ig_l2}{\dashed} & 0.223  & 0.084    & 0.111    & 0.024    & 0.214    & 0.06     & 0.315   & 0.171 & 0.223       & 0.074       & 0.295          & 0.048 \cr
  $\text{IG}_{\mu}$ \textcolor{ig_mean}{\dashed}& 0.211          & 0.112          & 0.106          & \textbf{0.022} & 0.179          & 0.063          & 0.317          & 0.174 & 0.231       & 0.067       & 0.279          & 0.064\cr
  $\text{G}\times\text{I}_{\ell_2}$ \textcolor{grad_input_l2}{\dashed} & 0.199    & 0.080     & 0.081    & 0.104    & 0.197    & 0.056    & 0.279   & 0.149 & 0.187       & 0.081       & 0.235          & 0.065\cr
  $\text{G}\times\text{I}_{\mu}$ \textcolor{grad_input_mean}{\dashed} & 0.207          & 0.073          & 0.087          & 0.098          & 0.213          & 0.054          & 0.285          & 0.145 & 0.191       & 0.078       & 0.237          & 0.065\cr
  Rollout \textcolor{rollout}{\dashed} & 0.074    & 0.270     & 0.076    & 0.102    & 0.09     & 0.185    & 0.108   & 0.292 & 0.076       & 0.179       & 0.152          & 0.147\cr
  Globenc \textcolor{norm2}{\dashed} & 0.174    & 0.125    & 0.118    & 0.053    & 0.207    & 0.117    & 0.251   & 0.178 & 0.161       & 0.119       & 0.24           & 0.072\cr
  ALTI \textcolor{alti}{\dashed} & \textbf{0.317} & \textbf{0.044} & \textbf{0.255} & \textbf{0.022} & \textbf{0.308} & \textbf{0.031} & \textbf{0.372} & \textbf{0.088} & \textbf{0.269}      & \textbf{0.057}       & \textbf{0.332}          & \textbf{0.034}\cr
  \bottomrule
\end{tabular}
}
\caption{Faithfulness results of the different interpretability methods for BERT, RoBERTa and DistilBERT on four different datasets. $\uparrow$ means a higher number indicates better performance, while $\downarrow$ means the opposite.}
\label{table:results_aopc}
\end{table*}
\begin{figure*}[!t]
    \centering
    \subfigure[]{\includegraphics[width=0.332\textwidth]{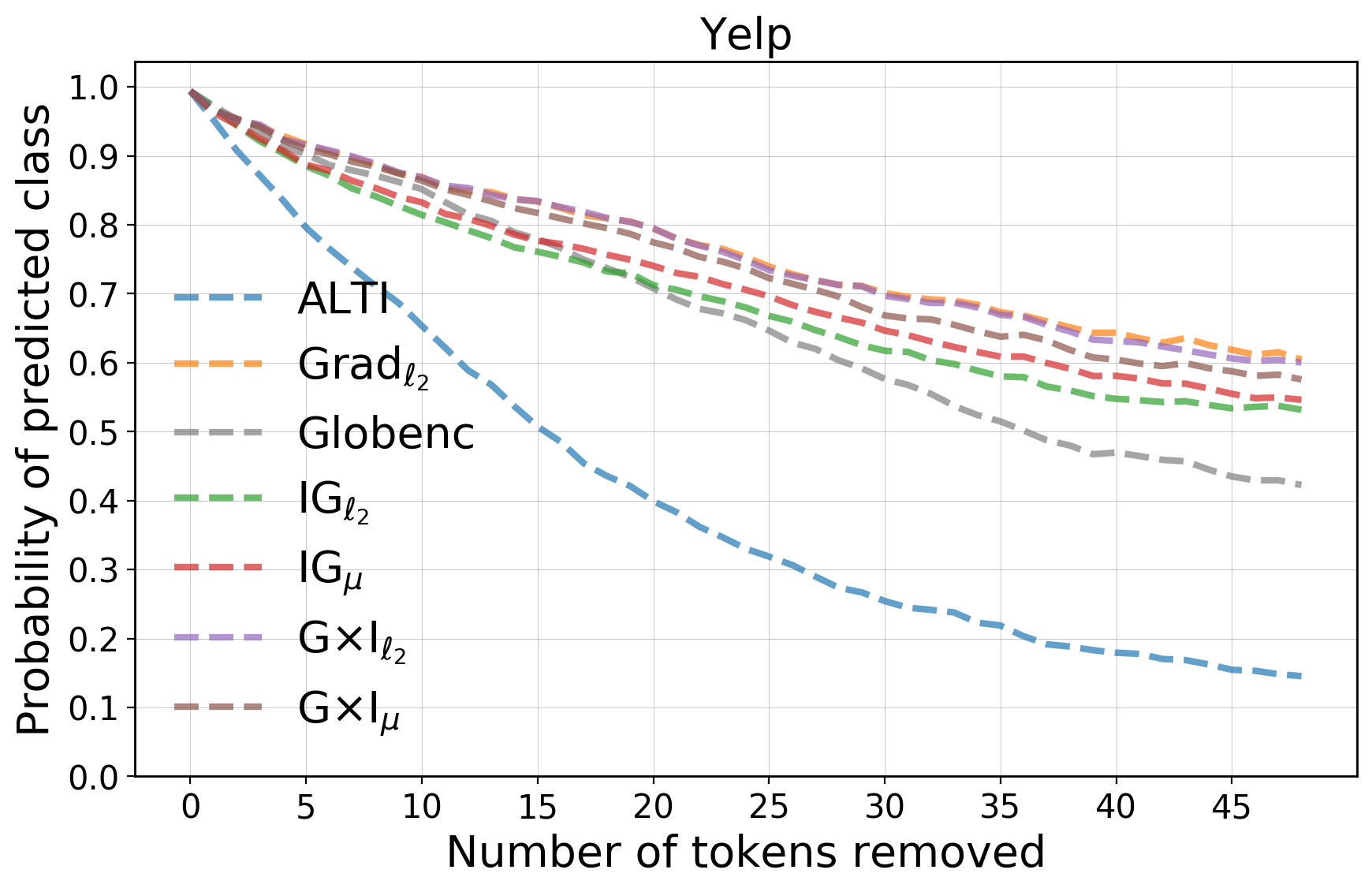}}
    \subfigure[]{\includegraphics[width=0.32\textwidth]{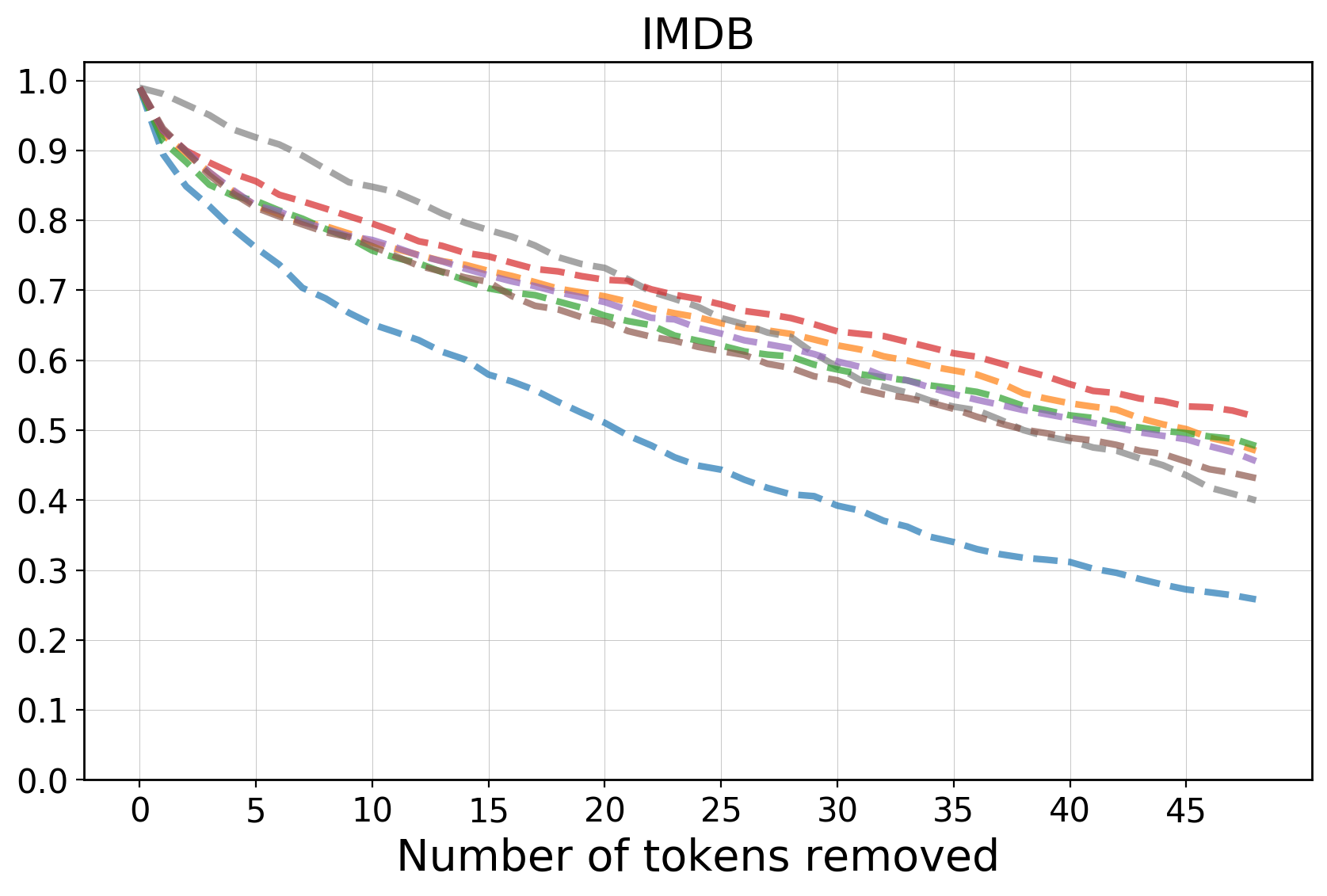}}
    \subfigure[]{\includegraphics[width=0.32\textwidth]{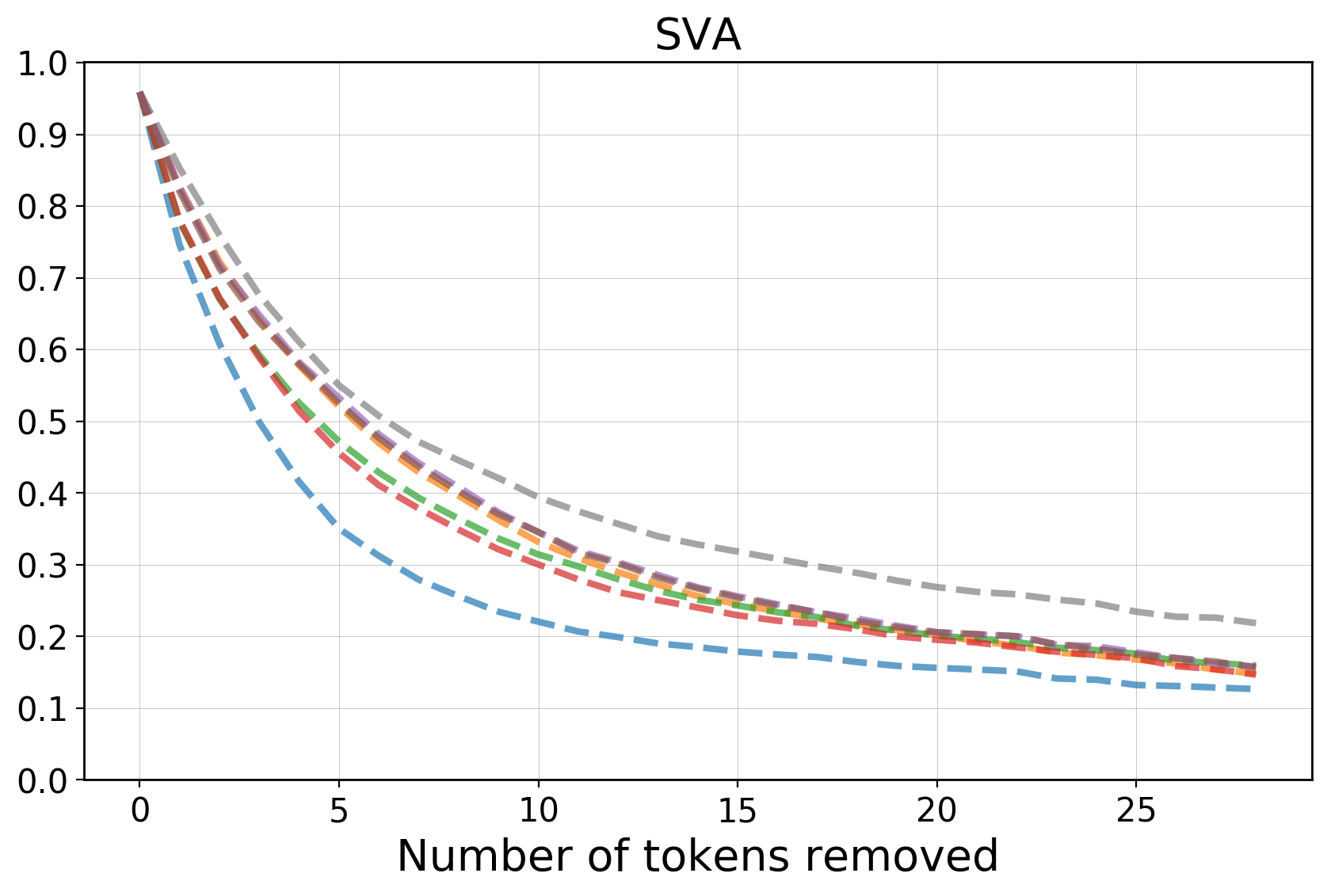}}
    \caption{Probability drop in BERT predictions when removing important tokens, obtained by different interpretability methods. We show results on three datasets.}
    \label{fig:prob_desc_3_datasets}
\end{figure*}

\section{Results}
In this section, we present quantitative and qualitative results comparing ALTI with other input attribution methods.
\subsection{Faithfulness Results}
In \Cref{table:results_aopc} we show comprehensiveness and sufficiency results for the three models and four datasets. It can be seen that across every different configuration, our proposed ALTI method outperforms other input attribution methods. Regarding comprehensiveness, datasets with short sentences like SST-2 and SVA (\Cref{fig:prob_desc_3_datasets} (c)) provide small differences between methods. This is expected since these datasets are simpler, and therefore, interpretations can more easily find the important tokens. However, for datasets containing longer inputs with multiple sentences, like IMDB and Yelp, ALTI clearly stands out. This can be observed in \Cref{fig:prob_desc_3_datasets} (a) and (b), where the probability drop in the model prediction is shown when removing one token at a time. We observe small differences in performance within gradient-based methods across datasets and models, with $\text{IG}_{\ell_2}$ performing the best on average among them, agreeing with the observations of \citet{atanasova-etal-2020-diagnostic}. However, ALTI outperforms $\text{IG}_{\ell_2}$ by 58\% on average in comprehensiveness, and by 38\% in sufficiency. Results of RoBERTa and DistilBERT on every dataset can be found in \Cref{apx:results}.

Previous research concluded that faithfulness results for evaluating different interpretability methods are task and model dependent \cite{bastings2021will, madsen_faithfulness}. Interestingly, although for the rest of the methods results vary across models and tasks, we observe ALTI repeatedly performs the best across different tasks and models.

In the qualitative examples in Tables \ref{tab:qualitative_examples_sst2_bert} and \ref{tab:qualitative_examples_yelp_bert} we can observe that gradient-based methods often miss the relevant tokens that drive the model's negative prediction. ALTI consistently assigns high relevance to spans of text that have a negative connotation, such as `depressing', `don't plan on returning' in \Cref{tab:qualitative_examples_yelp_bert}, or `food not good', `stomach aches' in \Cref{tab:qualitative_examples_sst2_bert}, as expected from a negative sentiment prediction. We observe that, as opposed to ALTI, gradient-based methods become less accurate with longer sequences. A very large example with a positive sentiment prediction can be found in \Cref{apx:qualitative_examples_sst2} \Cref{tab:long_imdb}, where ALTI accurately picks as important tokens those with positive meanings.

\begin{table*}[t!]
\small
\setlength{\tabcolsep}{0.2pt}
\begin{tabular}{ll}
\multicolumn{1}{l}{$\text{Grad}_{\ell_2}$}\\ 
 \mybox{color0}{\strut{um}} \mybox{color0}{\strut{.}} \mybox{color0}{\strut{it}} \mybox{color0}{\strut{'}} \mybox{color0}{\strut{s}} \mybox{color0}{\strut{okay}} \mybox{color0}{\strut{,}} \mybox{color0}{\strut{i}} \mybox{color0}{\strut{guess}} \mybox{color0}{\strut{.}} \mybox{color0}{\strut{they}} \mybox{color0}{\strut{have}} \mybox{color0}{\strut{food}} \mybox{color0}{\strut{at}} \mybox{color0}{\strut{decent}} \mybox{color0}{\strut{prices}} \mybox{color0}{\strut{,}} \mybox{color1}{\strut{but}} \mybox{color1}{\strut{the}} \mybox{color4}{\strut{isles}} \mybox{color1}{\strut{are}} \mybox{color0}{\strut{narrow}} \mybox{color0}{\strut{,}} \mybox{color1}{\strut{everything}} \mybox{color1}{\strut{needs}} \mybox{color0}{\strut{a}} \mybox{color0}{\strut{good}} \mybox{color0}{\strut{cleaning}} \mybox{color0}{\strut{and}}\\ \mybox{color1}{\strut{repainting}} \mybox{color0}{\strut{,}} \mybox{color0}{\strut{and}} \mybox{color0}{\strut{it}} \mybox{color0}{\strut{just}} \mybox{color0}{\strut{felt}} \mybox{color0}{\strut{dark}} \mybox{color0}{\strut{and}} \mybox{color1}{\strut{depressing}} \mybox{color0}{\strut{.}} \mybox{color0}{\strut{otherwise}} \mybox{color0}{\strut{it}} \mybox{color0}{\strut{'}} \mybox{color0}{\strut{s}} \mybox{color0}{\strut{all}} \mybox{color0}{\strut{right}} \mybox{color0}{\strut{,}} \mybox{color0}{\strut{but}} \mybox{color0}{\strut{i}} \mybox{color0}{\strut{don}} \mybox{color0}{\strut{'}} \mybox{color0}{\strut{t}} \mybox{color0}{\strut{plan}} \mybox{color0}{\strut{on}} \mybox{color1}{\strut{returning}} \mybox{color0}{\strut{here}} \mybox{color0}{\strut{.}}\\
\addlinespace
\multicolumn{1}{l}{IG$_{\ell_2}$}\\ 
 \mybox{color4}{\strut{um}} \mybox{color0}{\strut{.}} \mybox{color0}{\strut{it}} \mybox{color0}{\strut{'}} \mybox{color0}{\strut{s}} \mybox{color0}{\strut{okay}} \mybox{color0}{\strut{,}} \mybox{color0}{\strut{i}} \mybox{color0}{\strut{guess}} \mybox{color0}{\strut{.}} \mybox{color0}{\strut{they}} \mybox{color0}{\strut{have}} \mybox{color0}{\strut{food}} \mybox{color0}{\strut{at}} \mybox{color0}{\strut{decent}} \mybox{color0}{\strut{prices}} \mybox{color0}{\strut{,}} \mybox{color0}{\strut{but}} \mybox{color0}{\strut{the}} \mybox{color0}{\strut{isles}} \mybox{color0}{\strut{are}} \mybox{color0}{\strut{narrow}} \mybox{color0}{\strut{,}} \mybox{color0}{\strut{everything}} \mybox{color0}{\strut{needs}} \mybox{color0}{\strut{a}} \mybox{color0}{\strut{good}} \mybox{color0}{\strut{cleaning}} \mybox{color0}{\strut{and}}\\ \mybox{color0}{\strut{repainting}} \mybox{color0}{\strut{,}} \mybox{color0}{\strut{and}} \mybox{color0}{\strut{it}} \mybox{color0}{\strut{just}} \mybox{color0}{\strut{felt}} \mybox{color0}{\strut{dark}} \mybox{color0}{\strut{and}} \mybox{color0}{\strut{depressing}} \mybox{color0}{\strut{.}} \mybox{color0}{\strut{otherwise}} \mybox{color0}{\strut{it}} \mybox{color0}{\strut{'}} \mybox{color0}{\strut{s}} \mybox{color0}{\strut{all}} \mybox{color0}{\strut{right}} \mybox{color0}{\strut{,}} \mybox{color0}{\strut{but}} \mybox{color0}{\strut{i}} \mybox{color0}{\strut{don}} \mybox{color0}{\strut{'}} \mybox{color0}{\strut{t}} \mybox{color0}{\strut{plan}} \mybox{color0}{\strut{on}} \mybox{color0}{\strut{returning}} \mybox{color0}{\strut{here}} \mybox{color0}{\strut{.}}\\
\addlinespace
\multicolumn{1}{l}{ALTI}\\ 
 \mybox{color1}{\strut{um}} \mybox{color0}{\strut{.}} \mybox{color0}{\strut{it}} \mybox{color0}{\strut{'}} \mybox{color0}{\strut{s}} \mybox{color2}{\strut{okay}} \mybox{color0}{\strut{,}} \mybox{color0}{\strut{i}} \mybox{color3}{\strut{guess}} \mybox{color0}{\strut{.}} \mybox{color0}{\strut{they}} \mybox{color0}{\strut{have}} \mybox{color0}{\strut{food}} \mybox{color0}{\strut{at}} \mybox{color0}{\strut{decent}} \mybox{color0}{\strut{prices}} \mybox{color0}{\strut{,}} \mybox{color0}{\strut{but}} \mybox{color0}{\strut{the}} \mybox{color0}{\strut{isles}} \mybox{color0}{\strut{are}} \mybox{color0}{\strut{narrow}} \mybox{color0}{\strut{,}} \mybox{color0}{\strut{everything}} \mybox{color0}{\strut{needs}} \mybox{color0}{\strut{a}} \mybox{color0}{\strut{good}} \mybox{color0}{\strut{cleaning}} \mybox{color0}{\strut{and}}\\ \mybox{color1}{\strut{repainting}} \mybox{color0}{\strut{,}} \mybox{color0}{\strut{and}} \mybox{color0}{\strut{it}} \mybox{color0}{\strut{just}} \mybox{color1}{\strut{felt}} \mybox{color1}{\strut{dark}} \mybox{color0}{\strut{and}} \mybox{color2}{\strut{depressing}} \mybox{color0}{\strut{.}} \mybox{color2}{\strut{otherwise}} \mybox{color0}{\strut{it}} \mybox{color0}{\strut{'}} \mybox{color0}{\strut{s}} \mybox{color1}{\strut{all}} \mybox{color1}{\strut{right}} \mybox{color0}{\strut{,}} \mybox{color1}{\strut{but}} \mybox{color0}{\strut{i}} \mybox{color2}{\strut{don}} \mybox{color1}{\strut{'}} \mybox{color1}{\strut{t}} \mybox{color3}{\strut{plan}} \mybox{color1}{\strut{on}} \mybox{color4}{\strut{returning}} \mybox{color2}{\strut{here}} \mybox{color0}{\strut{.}}\\
\bottomrule
\end{tabular}
\caption{Saliency maps of BERT generated by two common gradient methods and by our proposed method, ALTI, for a \textbf{negative} sentiment predictions of Yelp dataset.}
\label{tab:qualitative_examples_yelp_bert}
\end{table*}

\subsection{Robustness Analysis}\label{sec:robustness_analysis}
We perform a study to investigate the robustness of different interpretability methods based on the \textit{implementation invariance} property defined by \citet{pmlr-v70-sundararajan17a}. Given a set of models with the same architecture, and trained with the same data, but only differing in their random weight initialization, it compares how different are the input attribution scores between the models, for the same interpretability method. If the predictions of the models are also identical, i.e. models are \textit{functionally equivalent}, we would expect input attributions also to be identical.
\begin{figure}[!t]
    \begin{centering}
    \includegraphics[width=0.44\textwidth]{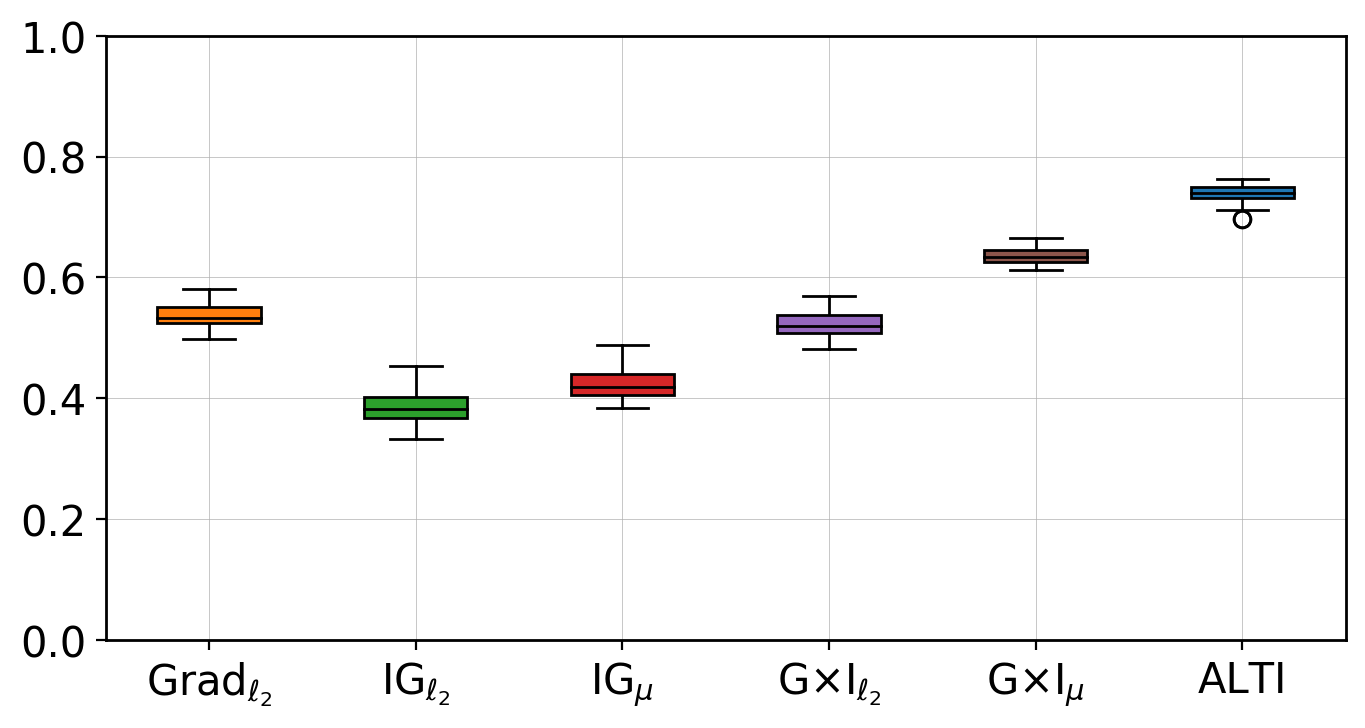}
    \caption{Jaccard-25\% similarity score between the interpretations of each method in 10 BERT's random seeds.}
    \label{fig:jaccard_seeds}
    \end{centering}
\end{figure}
\citet{zafar-etal-2021-lack} perform this test on two identical neural text classifiers $(i,j)$, differing in their random weight initialization. Since the vast majority of the predictions are the same for both models, they consider them to be \textit{almost functionally identical} models. Then, they measure the Jaccard similarity score between the top-$25\%$ tokens ranked based on their importance as specified by an input attribution method, for model $i$ and $j$. If the top-25\% tokens by the two attributions coincide, $\text{Jaccard-}25\%(i,j) = 1$. In case tokens don't overlap, $\text{Jaccard-}25\%(i,j) = 0$.

\begin{figure}[!t]
    \begin{centering}
    \includegraphics[width=0.44\textwidth]{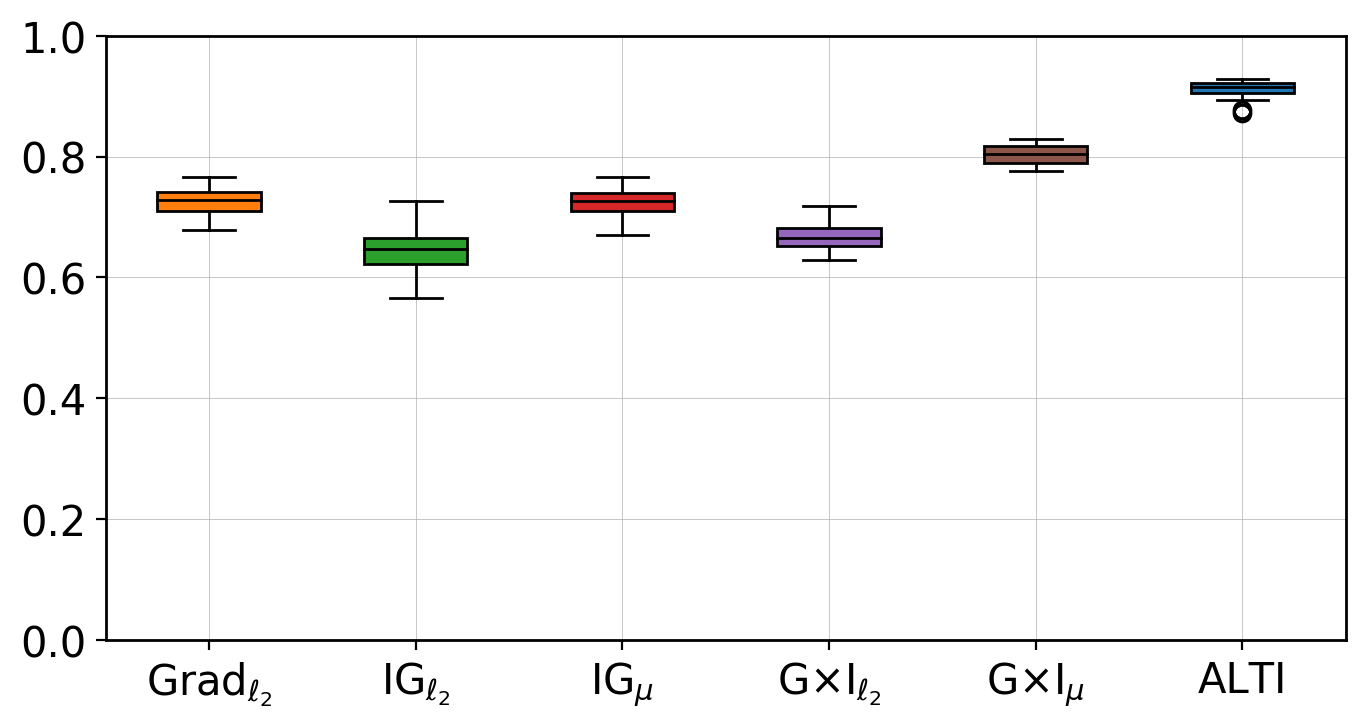}
    \caption{Spearman's rank correlation between the interpretations of each method in 10 BERT's random seeds.}
    \label{fig:corr_seeds}
    \end{centering}
\end{figure}

We perform the robustness test with 10 pre-trained BERT models from the MultiBERTs \cite{sellam2022the}, which only differ in their random weight initialization. For each interpretability method we compute Jaccard-25\% score between all the different pairs of models. In \Cref{fig:jaccard_seeds} we show the distribution of the obtained scores. We also compute the Spearman's rank correlation coefficient (\Cref{fig:corr_seeds}), which evaluates how well the relationship between the two ranked interpretations can be described using a monotonic function. We can observe that ALTI provides more homogeneous interpretations across identical models in terms of similarity and correlation, suggesting that it is a more robust interpretability method than gradient-based methods.

\subsection{Ablation Study}
We inspect the importance of the different components conforming ALTI.
\paragraph{Layer-wise token contributions.}\label{sec:alti_vs_kob_roll}
We compare the effect of our token contributions' measurement in Eq.~\ref{eq:distance} and Eq.~\ref{eq:contributions} against previous approach by \citet{kobayashi-etal-2021-incorporating} (Norms) described in Eq.~\ref{eq:kobayashi_contributions} by aggregating each type of contributions with the Rollout method. To isolate the influence of the norm choice, we use the $\ell_2$ in Eq.~\ref{eq:distance} and Eq.~\ref{eq:contributions} (ALTI $\ell_2$). Results in \Cref{tab:results_aopc_koba_vs_alti} show our proposed layer-wise contribution measurement largely improves previous approach.
\begin{table}[t]
\renewcommand{\arraystretch}{1}
\centering
\setlength{\tabcolsep}{4pt}
\resizebox{0.485\textwidth}{!}{%
\begin{tabular}{cccccccc}
\toprule
  & & \multicolumn{2}{c}{\bf{BERT}} & \multicolumn{2}{c}{\bf{RoBERTa}} & \multicolumn{2}{c}{\bf{DistilBERT}}\\
\cmidrule{3-4}\cmidrule{5-6}\cmidrule{7-8}
  \bf{Dataset} & \bf{Metric} & \bf{ALTI $\ell_2$}  & \bf{Norms} & \bf{ALTI $\ell_2$}  & \bf{Norms} & \bf{ALTI $\ell_2$}  & \bf{Norms}\\
  \midrule
\multirow{2}{*}{\bf{SST-2}} & \bf{Comp.$\uparrow$} & \textbf{0.3}   & 0.134 & \textbf{0.23}  & 0.168 & \textbf{0.292} & 0.157 \\
& \bf{Suff.$\downarrow$} & \textbf{0.045} & 0.18  & \textbf{0.075} & 0.135 & \textbf{0.051} & 0.152 \\
\midrule
\multirow{2}{*}{\bf{IMDB}} & \bf{Comp.} & \textbf{0.286} & 0.148 & \textbf{0.203} & 0.133 & \textbf{0.254} & 0.175 \\
& \bf{Suff.} & \textbf{0.031} & 0.115 & \textbf{0.086} & 0.162 & \textbf{0.053} & 0.104 \\
\midrule
\multirow{2}{*}{\bf{Yelp}} & \bf{Comp.} & \textbf{0.212} & 0.082 & \textbf{0.108} & 0.064 & \textbf{0.2}   & 0.093 \\
& \bf{Suff.} & \textbf{0.02}  & 0.057 & \textbf{0.104} & 0.18  & \textbf{0.031} & 0.114 \\
\midrule
\multirow{2}{*}{\bf{SVA}} & \bf{Comp.}  & \textbf{0.374} & 0.273 & \textbf{0.379} & 0.275 & \textbf{0.357} & 0.315 \\
& \bf{Suff.}  & \textbf{0.088} & 0.152 & \textbf{0.114} & 0.178 & \textbf{0.096} & 0.117 \\
\bottomrule
\end{tabular}
}
\caption{Faithfulness results of BERT, RoBERTa and DistilBERT comparing ALTI $\ell_2$ with the Norms approach.}
\label{tab:results_aopc_koba_vs_alti}
\end{table}

\begin{table}[!h]
\renewcommand{\arraystretch}{1}
\centering
\small
\setlength{\tabcolsep}{5pt}
\resizebox{0.485\textwidth}{!}{%
\begin{tabular}{cccccccc}
\toprule
  & & \multicolumn{2}{c}{\bf{BERT}} & \multicolumn{2}{c}{\bf{RoBERTa}} & \multicolumn{2}{c}{\bf{DistilBERT}}\\
\cmidrule{3-4}\cmidrule{5-6}\cmidrule{7-8}
  \bf{Dataset} & \bf{Metric} & \bf{$\ell_1$}  & \bf{$\ell_2$} & \bf{$\ell_1$}  & \bf{$\ell_2$} & \bf{$\ell_1$}  & \bf{$\ell_2$}\\
  \midrule
\multirow{2}{*}{\bf{SST-2}} & \bf{Comp.$\uparrow$} & \textbf{0.317}  & 0.3             & \textbf{0.269}       & 0.23          & \textbf{0.332}        & 0.292           \\
& \bf{Suff.$\downarrow$} & \textbf{0.044}  & 0.045           & \textbf{0.057}       & 0.075         & \textbf{0.034}        & 0.051           \\
\midrule
\multirow{2}{*}{\bf{IMDB}} & \bf{Comp.} & \textbf{0.308}  & 0.286           & \textbf{0.266}       & 0.203         & \textbf{0.304}        & 0.254           \\
& \bf{Suff.} & \textbf{0.031}  & \textbf{0.031}  & \textbf{0.05}        & 0.086         & \textbf{0.039}        & 0.053           \\
\midrule
\multirow{2}{*}{\bf{Yelp}} & \bf{Comp.} & \textbf{0.221}  & 0.212           & \textbf{0.138}       & 0.108         & \textbf{0.237}        & 0.2             \\
& \bf{Suff.} & \textbf{0.02}   & \textbf{0.02}   & \textbf{0.07}        & 0.104         & \textbf{0.017}        & 0.031           \\
\midrule
\multirow{2}{*}{\bf{SVA}} & \bf{Comp.} & 0.372           & \textbf{0.374}  & \textbf{0.39}        & 0.379         & \textbf{0.382}        & 0.357           \\
& \bf{Suff.}  & \textbf{0.088}  & \textbf{0.088}  & \textbf{0.109}       & 0.114         & \textbf{0.084}        & 0.096  \\
\bottomrule
\end{tabular}
}
\caption{Faithfulness results of BERT, RoBERTa and DistilBERT using $\ell_1$ and $\ell_2$ norms.}
\label{tab:results_aopc_l1_vs_l2}
\end{table}

\paragraph{Norm choice in ALTI.}\label{sec:l1_vs_l2} We also evaluate the election of the norm in our proposed approach. In \Cref{tab:results_aopc_l1_vs_l2} we show faithfulness results considering $\ell_1$ and $\ell_2$. In almost every setting $\ell_1$ outperforms $\ell_2$. Remarkably, the advantage of the $\ell_1$ is less noticeable on BERT, which we hypothesize is explained by the reduced anisotropy of its representations (\Cref{fig:mean_cosine_similarity}).


\subsection{Addition of Layer Norm 2}\label{sec:ln2_addition}
\begin{figure}[t]
    \begin{centering}
    \includegraphics[width=0.44\textwidth]{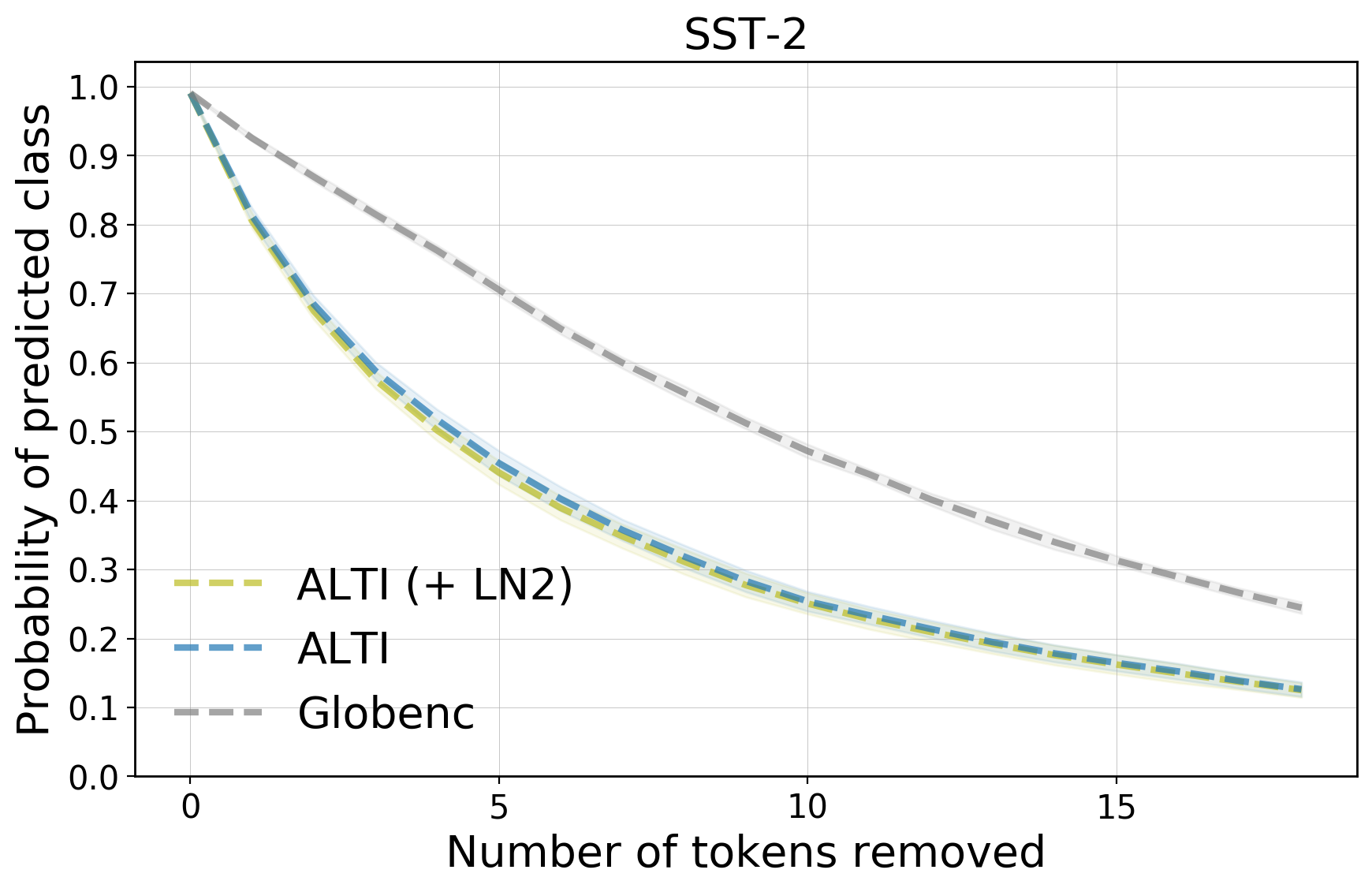}
    \caption{Probability drop in BERT predictions when removing important tokens, results show mean and SD in BERT across 10 seeds in SST-2 dataset.}
    \label{fig:prob_desc_alti_ln2_globenc}
    \end{centering}
\end{figure}

Concurrent work \cite{globenc} present Globenc method, which aggregates the contributions obtained by \cite{kobayashi-etal-2021-incorporating} in  Eq.~\ref{eq:kobayashi_contributions} with Attention Rollout method. Moreover, they add the layer normalization (LN2) of the Feed-forward module of the Transformer layer into their method. We evaluate the faithfulness of the interpretations provided by Globenc in \Cref{table:results_aopc}, and although it improves the Rollout baseline, is far from the results obtained with ALTI. We consider analyzing the influence of the second layer normalization by including it in ALTI method. The probability drop in SST-2 across 10 BERT seeds (\Cref{fig:prob_desc_alti_ln2_globenc}) shows the influence of LN2 is negligible. We observe similar patterns across models and datasets.

\section{Conclusions}
In this paper, we have presented ALTI, an input attribution method that quantifies the mixing of information in the Transformer. We have demonstrated that with accurate layer-wise token-to-token contribution measurements relying on $\ell_1$-based metrics, the interpretable attention decomposition of the attention block is a powerful tool when combined with the rollout method. Empirically, we show that ALTI outperforms every input attribution method we have experimented with in two common faithfulness metrics, while showing greater robustness. Overall, we believe this opens new possibilities for studying contextual information aggregation across the Transformer.

\section*{Limitations}
ALTI measures the amount of contextual information in each layer representation of the Transformer. From the influence of each input token to the last layer representation we extract input attributions for the model prediction. However, our method does not consider the classifier on top of the Transformer. Therefore, our proposed method doesn't provide explanations for each of the output classes, as opposed to gradient-based methods. We also underline that faithfulness in this work is evaluated via sufficiency and comprehensiveness metrics.

\section*{Ethical Considerations}
ALTI provides explanations about input attributions in the Transformer. By itself, we are not aware of any ethical implications of the methodology, which does not take into account any subjective priors. To prove its usefulness, we have used two different benchmarks, text classification, and subject-verb agreement. As far as we are concerned, these benchmarks have been used in the past without raising major ethical considerations. Therefore, we do not have any major issue to report in this section. 

\section{Acknowledgements}
We would like to thank the anonymous reviewers for their useful comments. Javier Ferrando and Gerard I. Gállego are supported by the Spanish Ministerio de Ciencia e Innovación through the project PID2019-107579RB-I00 / AEI / 10.13039/501100011033.

\bibliography{anthology,custom}
\bibliographystyle{acl_natbib}

\clearpage

\appendix

\section{Layer Normalization decomposition}\label{apx:ln_derivation}
Layer normalization acting over an input $\bm{u}$ can be defined as: $\text{LN}(\bm{u})=\frac{\bm{u}-\mu(\bm{u})}{\sigma(\bm{u})} \odot \mathbf{\gamma}+ \mathbf{\beta}$, where $\mu$ and $\sigma$ compute the mean and standard deviation of u, and $\gamma$ and $\beta$ refer to the element-wise transformation and bias respectively.
$\text{LN}(\bm{u})$ can be decomposed into $\frac{1}{\sigma(\bm{u})}\mathbf{L} \bm{u}+ \beta$, where $\mathbf{L}$ is a linear transformation:
\begin{equation*}
\resizebox{0.48\textwidth}{!}{$\displaystyle{
\mathbf{L}:=\left[ 
\begin{array}{cccc}
\gamma _{1} & 0 & \cdots  & 0 \\ 
0 & \gamma _{2} & \cdots  & 0 \\ 
\cdots  & \cdots  & \cdots  & \cdots  \\ 
0 & 0 & \cdots  & \gamma _{n}%
\end{array}%
\right] \left[ 
\begin{array}{cccc}
\frac{n-1}{n} & -\frac{1}{n} & \cdots  & -\frac{1}{n} \\ 
-\frac{1}{n} & \frac{n-1}{n} & \cdots  & -\frac{1}{n} \\ 
\cdots  & \cdots  & \cdots  & \cdots  \\ 
-\frac{1}{n} & -\frac{1}{n} & \cdots  & \frac{n-1}{n}%
\end{array}%
\right]}$}
\end{equation*}%

The linear map on the right subtracts the mean to the input vector, $\bm{u}' = \bm{u}-\mu(\bm{u})$. The left matrix performs the hadamard product with the layer normalization weights ($\bm{u}' \odot \gamma$).
  
\section{RoBERTa and DistilBERT Results}\label{apx:results}
\begin{table}[!h]
\renewcommand{\arraystretch}{0.9}
\centering
\small
\resizebox{0.49\textwidth}{!}{%
\begin{tabular}{lllllll}
\toprule
  & \multicolumn{2}{c}{\bf{IMDB}} & \multicolumn{2}{c}{\bf{Yelp}} & \multicolumn{2}{c}{\bf{SVA}}\\
\cmidrule{2-3}\cmidrule{4-5}\cmidrule{6-7}
  \bf{Methods} & \bf{Comp.$\uparrow$}  & \bf{Suff.$\downarrow$} & \bf{Comp.}  & \bf{Suff.} & \bf{Comp.}  & \bf{Suff.}\\
  \midrule
$\text{Grad}_{\ell_2}$                         & 0.266    & 0.054    & 0.111    & 0.073    & 0.338   & 0.107   \\
$\text{IG}_{\ell_2}$                       & 0.275    & 0.055    & 0.128    & 0.054    & 0.372   & 0.108   \\
$\text{IG}_{\mu}$                & 0.244    & 0.059    & 0.108    & 0.056    & 0.365   & 0.118   \\
$\text{G}\times\text{I}_{\ell_2}$              & 0.281    & 0.053    & 0.115    & 0.075    & 0.333   & 0.111   \\
$\text{G}\times\text{I}_{\mu}$ & 0.282    & 0.053    & 0.117    & 0.074    & 0.335   & 0.11    \\
Rollout                      & 0.183    & 0.089    & 0.092    & 0.114    & 0.241   & 0.179   \\
Globenc                        & 0.252    & 0.066    & 0.124    & 0.066    & 0.341   & 0.108   \\
ALTI                         & \bf{0.304}    & \bf{0.039}    & \bf{0.237}    & \bf{0.017}    & \bf{0.382}   & \bf{0.084}  \\
\bottomrule
\end{tabular}
}
\caption{Faithfulness results of the different interpretability methods for DistilBERT on IMDB, Yelp and SVA datasets. $\uparrow$ means a higher number indicates better performance, while $\downarrow$ means the opposite.}
\label{table:results_aopc_distilbert}
\end{table}

\begin{table}[!h]
\renewcommand{\arraystretch}{0.9}
\centering
\small
\resizebox{0.49\textwidth}{!}{%
\begin{tabular}{lllllll}
\toprule
  & \multicolumn{2}{c}{\bf{IMDB}} & \multicolumn{2}{c}{\bf{Yelp}} & \multicolumn{2}{c}{\bf{SVA}}\\
\cmidrule{2-3}\cmidrule{4-5}\cmidrule{6-7}
  \bf{Methods} & \bf{Comp.$\uparrow$}  & \bf{Suff.$\downarrow$} & \bf{Comp.}  & \bf{Suff.} & \bf{Comp.}  & \bf{Suff.}\\
  \midrule
$\text{Grad}_{\ell_2}$                         & 0.216    & 0.083    & 0.075    & 0.122    & 0.273   & 0.165   \\
$\text{IG}_{\ell_2}$                       & 0.2      & 0.084    & 0.087    & 0.11     & 0.363   & 0.163   \\
$\text{IG}_{\mu}$                & 0.215    & 0.083    & 0.102    & 0.094    & 0.351   & 0.161   \\
$\text{G}\times\text{I}_{\ell_2}$              & 0.183    & 0.109    & 0.066    & 0.149    & 0.28    & 0.169   \\
$\text{G}\times\text{I}_{\mu}$ & 0.225    & 0.08     & 0.083    & 0.12     & 0.27    & 0.167   \\
Rollout                      & 0.077    & 0.197    & 0.031    & 0.208    & 0.223   & 0.183   \\
Globenc                        & 0.154    & 0.086    & 0.065    & 0.12     & 0.305   & 0.17    \\
ALTI     & \bf{0.266}    & \bf{0.05}     & \bf{0.138}    & \bf{0.07}     & \bf{0.39}    & \bf{0.109} \\
\bottomrule
\end{tabular}
}
\caption{Faithfulness results of the different interpretability methods for RoBERTa on IMDB, Yelp and SVA datasets. $\uparrow$ means a higher number indicates better performance, while $\downarrow$ means the opposite.}
\label{table:results_aopc_roberta}
\end{table}

\section{Qualitative Examples}\label{apx:qualitative_examples_sst2}
\begin{table}[H]
\resizebox{0.49\textwidth}{!}{%
\setlength{\tabcolsep}{0.2pt}
\begin{tabular}{ll}
\multicolumn{1}{l}{$\text{Grad}_{\ell_2}$}\\ 
\mybox{color1}{\strut{friendly}} \mybox{color1}{\strut{staff}} \mybox{color0}{\strut{and}} \mybox{color2}{\strut{nice}} \mybox{color1}{\strut{selection}} \mybox{color1}{\strut{of}} \mybox{color1}{\strut{vegetarian}} \mybox{color0}{\strut{options}} \mybox{color2}{\strut{.}} \mybox{color4}{\strut{food}} \mybox{color1}{\strut{is}} \mybox{color2}{\strut{just}} \mybox{color1}{\strut{okay}} \mybox{color0}{\strut{,}}\\ \mybox{color1}{\strut{not}} \mybox{color1}{\strut{great}} \mybox{color0}{\strut{.}} \mybox{color0}{\strut{makes}} \mybox{color0}{\strut{me}} \mybox{color1}{\strut{wonder}} \mybox{color0}{\strut{why}} \mybox{color0}{\strut{everyone}} \mybox{color1}{\strut{likes}} \mybox{color0}{\strut{food}} \mybox{color0}{\strut{fight}} \mybox{color0}{\strut{so}} \mybox{color0}{\strut{much}} \mybox{color0}{\strut{.}} \\
\addlinespace
\multicolumn{1}{l}{G$\times$I$_{\ell_2}$}\\ 
\mybox{color1}{\strut{friendly}} \mybox{color2}{\strut{staff}} \mybox{color0}{\strut{and}} \mybox{color2}{\strut{nice}} \mybox{color1}{\strut{selection}} \mybox{color1}{\strut{of}} \mybox{color2}{\strut{vegetarian}} \mybox{color0}{\strut{options}} \mybox{color1}{\strut{.}} \mybox{color4}{\strut{food}} \mybox{color1}{\strut{is}} \mybox{color2}{\strut{just}} \mybox{color2}{\strut{okay}} \mybox{color0}{\strut{,}}\\ \mybox{color1}{\strut{not}} \mybox{color1}{\strut{great}} \mybox{color0}{\strut{.}} \mybox{color0}{\strut{makes}} \mybox{color0}{\strut{me}} \mybox{color2}{\strut{wonder}} \mybox{color1}{\strut{why}} \mybox{color1}{\strut{everyone}} \mybox{color1}{\strut{likes}} \mybox{color0}{\strut{food}} \mybox{color1}{\strut{fight}} \mybox{color0}{\strut{so}} \mybox{color0}{\strut{much}} \mybox{color0}{\strut{.}} \\
\addlinespace
\multicolumn{1}{l}{IG$_{\ell_2}$}\\ 
\mybox{color4}{\strut{friendly}} \mybox{color2}{\strut{staff}} \mybox{color1}{\strut{and}} \mybox{color1}{\strut{nice}} \mybox{color1}{\strut{selection}} \mybox{color0}{\strut{of}} \mybox{color1}{\strut{vegetarian}} \mybox{color0}{\strut{options}} \mybox{color0}{\strut{.}} \mybox{color2}{\strut{food}} \mybox{color1}{\strut{is}} \mybox{color1}{\strut{just}} \mybox{color1}{\strut{okay}} \mybox{color0}{\strut{,}}\\ \mybox{color1}{\strut{not}} \mybox{color2}{\strut{great}} \mybox{color0}{\strut{.}} \mybox{color0}{\strut{makes}} \mybox{color0}{\strut{me}} \mybox{color1}{\strut{wonder}} \mybox{color1}{\strut{why}} \mybox{color0}{\strut{everyone}} \mybox{color1}{\strut{likes}} \mybox{color0}{\strut{food}} \mybox{color0}{\strut{fight}} \mybox{color0}{\strut{so}} \mybox{color1}{\strut{much}} \mybox{color0}{\strut{.}}\\
\addlinespace
\multicolumn{1}{l}{ALTI}\\ 
\mybox{color1}{\strut{friendly}} \mybox{color1}{\strut{staff}} \mybox{color0}{\strut{and}} \mybox{color1}{\strut{nice}} \mybox{color1}{\strut{selection}} \mybox{color0}{\strut{of}} \mybox{color1}{\strut{vegetarian}} \mybox{color0}{\strut{options}} \mybox{color0}{\strut{.}} \mybox{color3}{\strut{food}} \mybox{color0}{\strut{is}} \mybox{color2}{\strut{just}} \mybox{color4}{\strut{okay}} \mybox{color1}{\strut{,}}\\ \mybox{color4}{\strut{not}} \mybox{color4}{\strut{great}} \mybox{color0}{\strut{.}} \mybox{color1}{\strut{makes}} \mybox{color0}{\strut{me}} \mybox{color1}{\strut{wonder}} \mybox{color2}{\strut{why}} \mybox{color0}{\strut{everyone}} \mybox{color0}{\strut{likes}} \mybox{color0}{\strut{food}} \mybox{color0}{\strut{fight}} \mybox{color0}{\strut{so}} \mybox{color0}{\strut{much}} \mybox{color0}{\strut{.}}\\
\bottomrule
\end{tabular}
}
\caption{Saliency maps of BERT generated by three common gradient methods and by our proposed method, ALTI, for a \textbf{negative} sentiment prediction example of Yelp dataset.} 
\end{table}

\begin{table*}[!t]
\begin{tabular}{ll}
\multicolumn{1}{l}{$\text{Grad}_{\ell_2}$}\\ 
 \mybox{color0}{\strut{low}} \mybox{color0}{\strut{budget}} \mybox{color0}{\strut{horror}} \mybox{color0}{\strut{movie}} \mybox{color0}{\strut{.}} \mybox{color0}{\strut{if}} \mybox{color0}{\strut{you}} \mybox{color0}{\strut{don}} \mybox{color0}{\strut{'}} \mybox{color0}{\strut{t}} \mybox{color0}{\strut{raise}} \mybox{color0}{\strut{your}} \mybox{color0}{\strut{expectations}} \mybox{color0}{\strut{too}} \mybox{color0}{\strut{high}} \mybox{color0}{\strut{,}} \mybox{color0}{\strut{you}} \mybox{color0}{\strut{'}} \mybox{color0}{\strut{ll}} \mybox{color0}{\strut{probably}} \mybox{color0}{\strut{enjoy}}\\ \mybox{color0}{\strut{this}} \mybox{color0}{\strut{little}} \mybox{color0}{\strut{flick}} \mybox{color0}{\strut{.}} \mybox{color0}{\strut{beginning}} \mybox{color0}{\strut{and}} \mybox{color0}{\strut{end}} \mybox{color0}{\strut{are}} \mybox{color0}{\strut{pretty}} \mybox{color0}{\strut{good}} \mybox{color0}{\strut{,}} \mybox{color0}{\strut{middle}} \mybox{color0}{\strut{drags}} \mybox{color0}{\strut{at}} \mybox{color0}{\strut{times}} \mybox{color0}{\strut{and}} \mybox{color0}{\strut{seems}} \mybox{color0}{\strut{to}} \mybox{color0}{\strut{go}} \mybox{color0}{\strut{nowhere}}\\ \mybox{color0}{\strut{for}} \mybox{color0}{\strut{long}} \mybox{color0}{\strut{periods}} \mybox{color0}{\strut{as}} \mybox{color0}{\strut{we}} \mybox{color0}{\strut{watch}} \mybox{color0}{\strut{the}} \mybox{color0}{\strut{goings}} \mybox{color0}{\strut{on}} \mybox{color0}{\strut{of}} \mybox{color0}{\strut{the}} \mybox{color0}{\strut{insane}} \mybox{color0}{\strut{that}} \mybox{color0}{\strut{add}} \mybox{color0}{\strut{atmosphere}} \mybox{color0}{\strut{but}} \mybox{color0}{\strut{do}} \mybox{color0}{\strut{not}} \mybox{color0}{\strut{advance}}\\ \mybox{color0}{\strut{the}} \mybox{color0}{\strut{plot}} \mybox{color0}{\strut{.}} \mybox{color0}{\strut{quite}} \mybox{color0}{\strut{a}} \mybox{color0}{\strut{bit}} \mybox{color0}{\strut{of}} \mybox{color0}{\strut{gore}} \mybox{color0}{\strut{.}} \mybox{color0}{\strut{i}} \mybox{color0}{\strut{enjoyed}} \mybox{color0}{\strut{bill}} \mybox{color0}{\strut{mcghee}} \mybox{color0}{\strut{'}} \mybox{color0}{\strut{s}} \mybox{color0}{\strut{performance}} \mybox{color0}{\strut{which}} \mybox{color0}{\strut{he}} \mybox{color0}{\strut{made}} \mybox{color0}{\strut{quite}} \mybox{color1}{\strut{believable}}\\ \mybox{color0}{\strut{for}} \mybox{color0}{\strut{such}} \mybox{color0}{\strut{a}} \mybox{color0}{\strut{low}} \mybox{color0}{\strut{budget}} \mybox{color0}{\strut{picture}} \mybox{color0}{\strut{,}} \mybox{color0}{\strut{he}} \mybox{color0}{\strut{managed}} \mybox{color0}{\strut{to}} \mybox{color0}{\strut{carry}} \mybox{color0}{\strut{the}} \mybox{color0}{\strut{movie}} \mybox{color0}{\strut{at}} \mybox{color0}{\strut{times}} \mybox{color0}{\strut{when}} \mybox{color0}{\strut{nothing}} \mybox{color0}{\strut{much}} \mybox{color0}{\strut{seemed}}\\ \mybox{color0}{\strut{to}} \mybox{color0}{\strut{be}} \mybox{color0}{\strut{happening}} \mybox{color0}{\strut{.}} \mybox{color0}{\strut{nurse}} \mybox{color0}{\strut{charlotte}} \mybox{color0}{\strut{beale}} \mybox{color0}{\strut{,}} \mybox{color0}{\strut{played}} \mybox{color0}{\strut{by}} \mybox{color0}{\strut{jesse}} \mybox{color0}{\strut{lee}} \mybox{color0}{\strut{,}} \mybox{color0}{\strut{played}} \mybox{color0}{\strut{her}} \mybox{color0}{\strut{character}} \mybox{color0}{\strut{well}} \mybox{color0}{\strut{so}} \mybox{color0}{\strut{be}}\\ \mybox{color0}{\strut{prepared}} \mybox{color0}{\strut{to}} \mybox{color0}{\strut{want}} \mybox{color0}{\strut{to}} \mybox{color0}{\strut{slap}} \mybox{color0}{\strut{her}} \mybox{color0}{\strut{toward}} \mybox{color0}{\strut{the}} \mybox{color0}{\strut{end}} \mybox{color0}{\strut{!}} \mybox{color0}{\strut{she}} \mybox{color0}{\strut{makes}} \mybox{color0}{\strut{some}} \mybox{color0}{\strut{really}} \mybox{color0}{\strut{stupid}} \mybox{color0}{\strut{mistakes}} \mybox{color0}{\strut{but}} \mybox{color0}{\strut{then}} \mybox{color0}{\strut{,}} \mybox{color0}{\strut{that}} \mybox{color0}{\strut{'}}\\ \mybox{color0}{\strut{s}} \mybox{color0}{\strut{what}} \mybox{color0}{\strut{makes}} \mybox{color0}{\strut{these}} \mybox{color0}{\strut{low}} \mybox{color0}{\strut{budget}} \mybox{color0}{\strut{movies}} \mybox{color0}{\strut{so}} \mybox{color0}{\strut{good}} \mybox{color0}{\strut{!}} \mybox{color0}{\strut{i}} \mybox{color0}{\strut{would}} \mybox{color0}{\strut{have}} \mybox{color0}{\strut{been}} \mybox{color0}{\strut{out}} \mybox{color0}{\strut{of}} \mybox{color0}{\strut{that}} \mybox{color0}{\strut{place}} \mybox{color0}{\strut{and}} \mybox{color0}{\strut{five}} \mybox{color0}{\strut{states}}\\ \mybox{color0}{\strut{away}} \mybox{color0}{\strut{long}} \mybox{color0}{\strut{before}} \mybox{color0}{\strut{she}} \mybox{color0}{\strut{even}} \mybox{color0}{\strut{considered}} \mybox{color0}{\strut{that}} \mybox{color0}{\strut{it}} \mybox{color0}{\strut{might}} \mybox{color0}{\strut{be}} \mybox{color0}{\strut{a}} \mybox{color0}{\strut{good}} \mybox{color0}{\strut{idea}} \mybox{color0}{\strut{to}} \mybox{color0}{\strut{leave}} \mybox{color0}{\strut{!}} \mybox{color0}{\strut{if}} \mybox{color0}{\strut{you}} \mybox{color1}{\strut{enjoy}} \mybox{color0}{\strut{this}} \mybox{color0}{\strut{movie}}\\ \mybox{color1}{\strut{,}} \mybox{color0}{\strut{try}} \mybox{color1}{\strut{committed}} \mybox{color0}{\strut{from}} \mybox{color0}{\strut{1988}} \mybox{color0}{\strut{which}} \mybox{color0}{\strut{is}} \mybox{color2}{\strut{basically}} \mybox{color2}{\strut{a}} \mybox{color4}{\strut{rip}} \mybox{color1}{\strut{off}} \mybox{color0}{\strut{of}} \mybox{color0}{\strut{this}} \mybox{color0}{\strut{movie}} \mybox{color0}{\strut{.}}\\
\addlinespace
\multicolumn{1}{l}{G$\times$I$_{\ell_2}$}\\ 
 \mybox{color0}{\strut{low}} \mybox{color0}{\strut{budget}} \mybox{color0}{\strut{horror}} \mybox{color0}{\strut{movie}} \mybox{color0}{\strut{.}} \mybox{color0}{\strut{if}} \mybox{color0}{\strut{you}} \mybox{color0}{\strut{don}} \mybox{color0}{\strut{'}} \mybox{color0}{\strut{t}} \mybox{color0}{\strut{raise}} \mybox{color0}{\strut{your}} \mybox{color0}{\strut{expectations}} \mybox{color0}{\strut{too}} \mybox{color0}{\strut{high}} \mybox{color0}{\strut{,}} \mybox{color0}{\strut{you}} \mybox{color0}{\strut{'}} \mybox{color0}{\strut{ll}} \mybox{color0}{\strut{probably}} \mybox{color0}{\strut{enjoy}}\\ \mybox{color0}{\strut{this}} \mybox{color0}{\strut{little}} \mybox{color0}{\strut{flick}} \mybox{color0}{\strut{.}} \mybox{color0}{\strut{beginning}} \mybox{color0}{\strut{and}} \mybox{color0}{\strut{end}} \mybox{color0}{\strut{are}} \mybox{color0}{\strut{pretty}} \mybox{color0}{\strut{good}} \mybox{color0}{\strut{,}} \mybox{color0}{\strut{middle}} \mybox{color0}{\strut{drags}} \mybox{color0}{\strut{at}} \mybox{color0}{\strut{times}} \mybox{color0}{\strut{and}} \mybox{color0}{\strut{seems}} \mybox{color0}{\strut{to}} \mybox{color0}{\strut{go}} \mybox{color0}{\strut{nowhere}}\\ \mybox{color0}{\strut{for}} \mybox{color0}{\strut{long}} \mybox{color0}{\strut{periods}} \mybox{color0}{\strut{as}} \mybox{color0}{\strut{we}} \mybox{color0}{\strut{watch}} \mybox{color0}{\strut{the}} \mybox{color0}{\strut{goings}} \mybox{color0}{\strut{on}} \mybox{color0}{\strut{of}} \mybox{color0}{\strut{the}} \mybox{color0}{\strut{insane}} \mybox{color0}{\strut{that}} \mybox{color0}{\strut{add}} \mybox{color0}{\strut{atmosphere}} \mybox{color0}{\strut{but}} \mybox{color0}{\strut{do}} \mybox{color0}{\strut{not}} \mybox{color0}{\strut{advance}}\\ \mybox{color0}{\strut{the}} \mybox{color0}{\strut{plot}} \mybox{color0}{\strut{.}} \mybox{color0}{\strut{quite}} \mybox{color0}{\strut{a}} \mybox{color0}{\strut{bit}} \mybox{color0}{\strut{of}} \mybox{color0}{\strut{gore}} \mybox{color0}{\strut{.}} \mybox{color0}{\strut{i}} \mybox{color0}{\strut{enjoyed}} \mybox{color0}{\strut{bill}} \mybox{color0}{\strut{mcghee}} \mybox{color0}{\strut{'}} \mybox{color0}{\strut{s}} \mybox{color0}{\strut{performance}} \mybox{color0}{\strut{which}} \mybox{color0}{\strut{he}} \mybox{color0}{\strut{made}} \mybox{color0}{\strut{quite}} \mybox{color1}{\strut{believable}}\\ \mybox{color0}{\strut{for}} \mybox{color0}{\strut{such}} \mybox{color0}{\strut{a}} \mybox{color0}{\strut{low}} \mybox{color0}{\strut{budget}} \mybox{color0}{\strut{picture}} \mybox{color0}{\strut{,}} \mybox{color0}{\strut{he}} \mybox{color0}{\strut{managed}} \mybox{color0}{\strut{to}} \mybox{color0}{\strut{carry}} \mybox{color0}{\strut{the}} \mybox{color0}{\strut{movie}} \mybox{color0}{\strut{at}} \mybox{color0}{\strut{times}} \mybox{color0}{\strut{when}} \mybox{color0}{\strut{nothing}} \mybox{color0}{\strut{much}} \mybox{color0}{\strut{seemed}}\\ \mybox{color0}{\strut{to}} \mybox{color0}{\strut{be}} \mybox{color0}{\strut{happening}} \mybox{color0}{\strut{.}} \mybox{color0}{\strut{nurse}} \mybox{color0}{\strut{charlotte}} \mybox{color0}{\strut{beale}} \mybox{color0}{\strut{,}} \mybox{color0}{\strut{played}} \mybox{color0}{\strut{by}} \mybox{color0}{\strut{jesse}} \mybox{color0}{\strut{lee}} \mybox{color0}{\strut{,}} \mybox{color0}{\strut{played}} \mybox{color0}{\strut{her}} \mybox{color0}{\strut{character}} \mybox{color0}{\strut{well}} \mybox{color0}{\strut{so}} \mybox{color0}{\strut{be}}\\ \mybox{color0}{\strut{prepared}} \mybox{color0}{\strut{to}} \mybox{color0}{\strut{want}} \mybox{color0}{\strut{to}} \mybox{color0}{\strut{slap}} \mybox{color0}{\strut{her}} \mybox{color0}{\strut{toward}} \mybox{color0}{\strut{the}} \mybox{color0}{\strut{end}} \mybox{color0}{\strut{!}} \mybox{color0}{\strut{she}} \mybox{color0}{\strut{makes}} \mybox{color0}{\strut{some}} \mybox{color0}{\strut{really}} \mybox{color0}{\strut{stupid}} \mybox{color0}{\strut{mistakes}} \mybox{color0}{\strut{but}} \mybox{color0}{\strut{then}} \mybox{color0}{\strut{,}} \mybox{color0}{\strut{that}}\\ \mybox{color0}{\strut{'}} \mybox{color0}{\strut{s}} \mybox{color0}{\strut{what}} \mybox{color0}{\strut{makes}} \mybox{color0}{\strut{these}} \mybox{color0}{\strut{low}} \mybox{color0}{\strut{budget}} \mybox{color0}{\strut{movies}} \mybox{color0}{\strut{so}} \mybox{color0}{\strut{good}} \mybox{color0}{\strut{!}} \mybox{color0}{\strut{i}} \mybox{color0}{\strut{would}} \mybox{color0}{\strut{have}} \mybox{color0}{\strut{been}} \mybox{color0}{\strut{out}} \mybox{color0}{\strut{of}} \mybox{color0}{\strut{that}} \mybox{color0}{\strut{place}} \mybox{color0}{\strut{and}} \mybox{color0}{\strut{five}} \mybox{color0}{\strut{states}}\\ \mybox{color0}{\strut{away}} \mybox{color0}{\strut{long}} \mybox{color0}{\strut{before}} \mybox{color0}{\strut{she}} \mybox{color0}{\strut{even}} \mybox{color0}{\strut{considered}} \mybox{color0}{\strut{that}} \mybox{color0}{\strut{it}} \mybox{color0}{\strut{might}} \mybox{color0}{\strut{be}} \mybox{color0}{\strut{a}} \mybox{color0}{\strut{good}} \mybox{color0}{\strut{idea}} \mybox{color0}{\strut{to}} \mybox{color0}{\strut{leave}} \mybox{color0}{\strut{!}} \mybox{color0}{\strut{if}} \mybox{color0}{\strut{you}} \mybox{color0}{\strut{enjoy}} \mybox{color0}{\strut{this}} \mybox{color0}{\strut{movie}}\\ \mybox{color0}{\strut{,}} \mybox{color0}{\strut{try}} \mybox{color1}{\strut{committed}} \mybox{color0}{\strut{from}} \mybox{color0}{\strut{1988}} \mybox{color0}{\strut{which}} \mybox{color0}{\strut{is}} \mybox{color2}{\strut{basically}} \mybox{color1}{\strut{a}} \mybox{color4}{\strut{rip}} \mybox{color1}{\strut{off}} \mybox{color0}{\strut{of}} \mybox{color0}{\strut{this}} \mybox{color0}{\strut{movie}} \mybox{color0}{\strut{.}}\\
\addlinespace
\multicolumn{1}{l}{IG$_{\ell_2}$}\\ 
 \mybox{color0}{\strut{low}} \mybox{color0}{\strut{budget}} \mybox{color0}{\strut{horror}} \mybox{color0}{\strut{movie}} \mybox{color0}{\strut{.}} \mybox{color0}{\strut{if}} \mybox{color0}{\strut{you}} \mybox{color0}{\strut{don}} \mybox{color0}{\strut{'}} \mybox{color0}{\strut{t}} \mybox{color0}{\strut{raise}} \mybox{color0}{\strut{your}} \mybox{color0}{\strut{expectations}} \mybox{color0}{\strut{too}} \mybox{color0}{\strut{high}} \mybox{color0}{\strut{,}} \mybox{color0}{\strut{you}} \mybox{color0}{\strut{'}} \mybox{color0}{\strut{ll}} \mybox{color1}{\strut{probably}} \mybox{color2}{\strut{enjoy}}\\ \mybox{color0}{\strut{this}} \mybox{color0}{\strut{little}} \mybox{color1}{\strut{flick}} \mybox{color0}{\strut{.}} \mybox{color0}{\strut{beginning}} \mybox{color0}{\strut{and}} \mybox{color1}{\strut{end}} \mybox{color0}{\strut{are}} \mybox{color1}{\strut{pretty}} \mybox{color1}{\strut{good}} \mybox{color0}{\strut{,}} \mybox{color0}{\strut{middle}} \mybox{color1}{\strut{drags}} \mybox{color0}{\strut{at}} \mybox{color1}{\strut{times}} \mybox{color0}{\strut{and}} \mybox{color0}{\strut{seems}} \mybox{color0}{\strut{to}} \mybox{color0}{\strut{go}} \mybox{color1}{\strut{nowhere}}\\ \mybox{color0}{\strut{for}} \mybox{color0}{\strut{long}} \mybox{color0}{\strut{periods}} \mybox{color0}{\strut{as}} \mybox{color1}{\strut{we}} \mybox{color0}{\strut{watch}} \mybox{color0}{\strut{the}} \mybox{color1}{\strut{goings}} \mybox{color0}{\strut{on}} \mybox{color0}{\strut{of}} \mybox{color0}{\strut{the}} \mybox{color0}{\strut{insane}} \mybox{color0}{\strut{that}} \mybox{color1}{\strut{add}} \mybox{color1}{\strut{atmosphere}} \mybox{color0}{\strut{but}} \mybox{color1}{\strut{do}} \mybox{color0}{\strut{not}} \mybox{color1}{\strut{advance}}\\ \mybox{color0}{\strut{the}} \mybox{color1}{\strut{plot}} \mybox{color0}{\strut{.}} \mybox{color0}{\strut{quite}} \mybox{color0}{\strut{a}} \mybox{color0}{\strut{bit}} \mybox{color0}{\strut{of}} \mybox{color0}{\strut{gore}} \mybox{color0}{\strut{.}} \mybox{color0}{\strut{i}} \mybox{color4}{\strut{enjoyed}} \mybox{color0}{\strut{bill}} \mybox{color2}{\strut{mcghee}} \mybox{color0}{\strut{'}} \mybox{color0}{\strut{s}} \mybox{color0}{\strut{performance}} \mybox{color0}{\strut{which}} \mybox{color0}{\strut{he}} \mybox{color0}{\strut{made}} \mybox{color1}{\strut{quite}} \mybox{color4}{\strut{believable}}\\ \mybox{color0}{\strut{for}} \mybox{color0}{\strut{such}} \mybox{color0}{\strut{a}} \mybox{color0}{\strut{low}} \mybox{color0}{\strut{budget}} \mybox{color0}{\strut{picture}} \mybox{color0}{\strut{,}} \mybox{color0}{\strut{he}} \mybox{color1}{\strut{managed}} \mybox{color0}{\strut{to}} \mybox{color1}{\strut{carry}} \mybox{color0}{\strut{the}} \mybox{color1}{\strut{movie}} \mybox{color0}{\strut{at}} \mybox{color1}{\strut{times}} \mybox{color1}{\strut{when}} \mybox{color1}{\strut{nothing}} \mybox{color0}{\strut{much}} \mybox{color1}{\strut{seemed}}\\ \mybox{color0}{\strut{to}} \mybox{color0}{\strut{be}} \mybox{color1}{\strut{happening}} \mybox{color0}{\strut{.}} \mybox{color0}{\strut{nurse}} \mybox{color1}{\strut{charlotte}} \mybox{color1}{\strut{beale}} \mybox{color0}{\strut{,}} \mybox{color1}{\strut{played}} \mybox{color0}{\strut{by}} \mybox{color1}{\strut{jesse}} \mybox{color0}{\strut{lee}} \mybox{color0}{\strut{,}} \mybox{color1}{\strut{played}} \mybox{color0}{\strut{her}} \mybox{color1}{\strut{character}} \mybox{color1}{\strut{well}} \mybox{color0}{\strut{so}} \mybox{color1}{\strut{be}}\\ \mybox{color1}{\strut{prepared}} \mybox{color0}{\strut{to}} \mybox{color1}{\strut{want}} \mybox{color1}{\strut{to}} \mybox{color1}{\strut{slap}} \mybox{color1}{\strut{her}} \mybox{color1}{\strut{toward}} \mybox{color0}{\strut{the}} \mybox{color2}{\strut{end}} \mybox{color1}{\strut{!}} \mybox{color1}{\strut{she}} \mybox{color1}{\strut{makes}} \mybox{color1}{\strut{some}} \mybox{color2}{\strut{really}} \mybox{color1}{\strut{stupid}} \mybox{color1}{\strut{mistakes}} \mybox{color0}{\strut{but}} \mybox{color0}{\strut{then}} \mybox{color0}{\strut{,}} \mybox{color0}{\strut{that}}\\ \mybox{color0}{\strut{'}} \mybox{color0}{\strut{s}} \mybox{color0}{\strut{what}} \mybox{color1}{\strut{makes}} \mybox{color0}{\strut{these}} \mybox{color0}{\strut{low}} \mybox{color0}{\strut{budget}} \mybox{color0}{\strut{movies}} \mybox{color0}{\strut{so}} \mybox{color0}{\strut{good}} \mybox{color0}{\strut{!}} \mybox{color0}{\strut{i}} \mybox{color1}{\strut{would}} \mybox{color0}{\strut{have}} \mybox{color0}{\strut{been}} \mybox{color0}{\strut{out}} \mybox{color0}{\strut{of}} \mybox{color0}{\strut{that}} \mybox{color1}{\strut{place}} \mybox{color0}{\strut{and}} \mybox{color0}{\strut{five}} \mybox{color1}{\strut{states}}\\ \mybox{color0}{\strut{away}} \mybox{color0}{\strut{long}} \mybox{color0}{\strut{before}} \mybox{color0}{\strut{she}} \mybox{color0}{\strut{even}} \mybox{color0}{\strut{considered}} \mybox{color0}{\strut{that}} \mybox{color0}{\strut{it}} \mybox{color0}{\strut{might}} \mybox{color0}{\strut{be}} \mybox{color0}{\strut{a}} \mybox{color0}{\strut{good}} \mybox{color0}{\strut{idea}} \mybox{color0}{\strut{to}} \mybox{color0}{\strut{leave}} \mybox{color0}{\strut{!}} \mybox{color1}{\strut{if}} \mybox{color0}{\strut{you}} \mybox{color4}{\strut{enjoy}} \mybox{color0}{\strut{this}} \mybox{color0}{\strut{movie}}\\ \mybox{color0}{\strut{,}} \mybox{color2}{\strut{try}} \mybox{color1}{\strut{committed}} \mybox{color0}{\strut{from}} \mybox{color1}{\strut{1988}} \mybox{color0}{\strut{which}} \mybox{color0}{\strut{is}} \mybox{color1}{\strut{basically}} \mybox{color0}{\strut{a}} \mybox{color2}{\strut{rip}} \mybox{color1}{\strut{off}} \mybox{color0}{\strut{of}} \mybox{color0}{\strut{this}} \mybox{color0}{\strut{movie}} \mybox{color0}{\strut{.}}\\
\addlinespace
\multicolumn{1}{l}{ALTI}\\ 
 \mybox{color2}{\strut{low}} \mybox{color3}{\strut{budget}} \mybox{color2}{\strut{horror}} \mybox{color2}{\strut{movie}} \mybox{color0}{\strut{.}} \mybox{color0}{\strut{if}} \mybox{color0}{\strut{you}} \mybox{color0}{\strut{don}} \mybox{color0}{\strut{'}} \mybox{color0}{\strut{t}} \mybox{color0}{\strut{raise}} \mybox{color0}{\strut{your}} \mybox{color0}{\strut{expectations}} \mybox{color0}{\strut{too}} \mybox{color0}{\strut{high}} \mybox{color0}{\strut{,}} \mybox{color0}{\strut{you}} \mybox{color0}{\strut{'}} \mybox{color1}{\strut{ll}} \mybox{color0}{\strut{probably}} \mybox{color2}{\strut{enjoy}}\\ \mybox{color3}{\strut{this}} \mybox{color4}{\strut{little}} \mybox{color4}{\strut{flick}} \mybox{color0}{\strut{.}} \mybox{color1}{\strut{beginning}} \mybox{color0}{\strut{and}} \mybox{color0}{\strut{end}} \mybox{color1}{\strut{are}} \mybox{color2}{\strut{pretty}} \mybox{color2}{\strut{good}} \mybox{color0}{\strut{,}} \mybox{color0}{\strut{middle}} \mybox{color0}{\strut{drags}} \mybox{color0}{\strut{at}} \mybox{color0}{\strut{times}} \mybox{color0}{\strut{and}} \mybox{color0}{\strut{seems}} \mybox{color0}{\strut{to}} \mybox{color0}{\strut{go}} \mybox{color0}{\strut{nowhere}}\\ \mybox{color0}{\strut{for}} \mybox{color0}{\strut{long}} \mybox{color0}{\strut{periods}} \mybox{color0}{\strut{as}} \mybox{color1}{\strut{we}} \mybox{color1}{\strut{watch}} \mybox{color0}{\strut{the}} \mybox{color2}{\strut{goings}} \mybox{color0}{\strut{on}} \mybox{color0}{\strut{of}} \mybox{color0}{\strut{the}} \mybox{color1}{\strut{insane}} \mybox{color0}{\strut{that}} \mybox{color1}{\strut{add}} \mybox{color1}{\strut{atmosphere}} \mybox{color0}{\strut{but}} \mybox{color0}{\strut{do}} \mybox{color0}{\strut{not}} \mybox{color0}{\strut{advance}}\\ \mybox{color0}{\strut{the}} \mybox{color0}{\strut{plot}} \mybox{color0}{\strut{.}} \mybox{color1}{\strut{quite}} \mybox{color0}{\strut{a}} \mybox{color0}{\strut{bit}} \mybox{color0}{\strut{of}} \mybox{color1}{\strut{gore}} \mybox{color0}{\strut{.}} \mybox{color1}{\strut{i}} \mybox{color2}{\strut{enjoyed}} \mybox{color0}{\strut{bill}} \mybox{color0}{\strut{mcghee}} \mybox{color0}{\strut{'}} \mybox{color0}{\strut{s}} \mybox{color1}{\strut{performance}} \mybox{color0}{\strut{which}} \mybox{color0}{\strut{he}} \mybox{color0}{\strut{made}} \mybox{color1}{\strut{quite}} \mybox{color4}{\strut{believable}}\\ \mybox{color0}{\strut{for}} \mybox{color0}{\strut{such}} \mybox{color0}{\strut{a}} \mybox{color0}{\strut{low}} \mybox{color0}{\strut{budget}} \mybox{color0}{\strut{picture}} \mybox{color0}{\strut{,}} \mybox{color0}{\strut{he}} \mybox{color0}{\strut{managed}} \mybox{color0}{\strut{to}} \mybox{color0}{\strut{carry}} \mybox{color0}{\strut{the}} \mybox{color1}{\strut{movie}} \mybox{color0}{\strut{at}} \mybox{color0}{\strut{times}} \mybox{color0}{\strut{when}} \mybox{color0}{\strut{nothing}} \mybox{color0}{\strut{much}} \mybox{color0}{\strut{seemed}}\\ \mybox{color0}{\strut{to}} \mybox{color0}{\strut{be}} \mybox{color0}{\strut{happening}} \mybox{color0}{\strut{.}} \mybox{color0}{\strut{nurse}} \mybox{color0}{\strut{charlotte}} \mybox{color0}{\strut{beale}} \mybox{color0}{\strut{,}} \mybox{color0}{\strut{played}} \mybox{color0}{\strut{by}} \mybox{color0}{\strut{jesse}} \mybox{color0}{\strut{lee}} \mybox{color0}{\strut{,}} \mybox{color0}{\strut{played}} \mybox{color0}{\strut{her}} \mybox{color1}{\strut{character}} \mybox{color0}{\strut{well}} \mybox{color0}{\strut{so}} \mybox{color0}{\strut{be}}\\ \mybox{color0}{\strut{prepared}} \mybox{color0}{\strut{to}} \mybox{color0}{\strut{want}} \mybox{color0}{\strut{to}} \mybox{color0}{\strut{slap}} \mybox{color0}{\strut{her}} \mybox{color0}{\strut{toward}} \mybox{color0}{\strut{the}} \mybox{color0}{\strut{end}} \mybox{color0}{\strut{!}} \mybox{color0}{\strut{she}} \mybox{color0}{\strut{makes}} \mybox{color0}{\strut{some}} \mybox{color0}{\strut{really}} \mybox{color0}{\strut{stupid}} \mybox{color0}{\strut{mistakes}} \mybox{color0}{\strut{but}} \mybox{color0}{\strut{then}} \mybox{color0}{\strut{,}} \mybox{color0}{\strut{that}}\\ \mybox{color0}{\strut{'}} \mybox{color0}{\strut{s}} \mybox{color1}{\strut{what}} \mybox{color2}{\strut{makes}} \mybox{color3}{\strut{these}} \mybox{color1}{\strut{low}} \mybox{color1}{\strut{budget}} \mybox{color2}{\strut{movies}} \mybox{color1}{\strut{so}} \mybox{color3}{\strut{good}} \mybox{color1}{\strut{!}} \mybox{color0}{\strut{i}} \mybox{color0}{\strut{would}} \mybox{color0}{\strut{have}} \mybox{color0}{\strut{been}} \mybox{color0}{\strut{out}} \mybox{color0}{\strut{of}} \mybox{color0}{\strut{that}} \mybox{color0}{\strut{place}} \mybox{color0}{\strut{and}} \mybox{color0}{\strut{five}} \mybox{color0}{\strut{states}}\\ \mybox{color0}{\strut{away}} \mybox{color0}{\strut{long}} \mybox{color0}{\strut{before}} \mybox{color0}{\strut{she}} \mybox{color0}{\strut{even}} \mybox{color0}{\strut{considered}} \mybox{color0}{\strut{that}} \mybox{color0}{\strut{it}} \mybox{color0}{\strut{might}} \mybox{color0}{\strut{be}} \mybox{color0}{\strut{a}} \mybox{color0}{\strut{good}} \mybox{color0}{\strut{idea}} \mybox{color0}{\strut{to}} \mybox{color0}{\strut{leave}} \mybox{color0}{\strut{!}} \mybox{color1}{\strut{if}} \mybox{color1}{\strut{you}} \mybox{color3}{\strut{enjoy}} \mybox{color1}{\strut{this}} \mybox{color1}{\strut{movie}}\\ \mybox{color0}{\strut{,}} \mybox{color0}{\strut{try}} \mybox{color0}{\strut{committed}} \mybox{color0}{\strut{from}} \mybox{color0}{\strut{1988}} \mybox{color0}{\strut{which}} \mybox{color0}{\strut{is}} \mybox{color0}{\strut{basically}} \mybox{color0}{\strut{a}} \mybox{color0}{\strut{rip}} \mybox{color0}{\strut{off}} \mybox{color0}{\strut{of}} \mybox{color0}{\strut{this}} \mybox{color1}{\strut{movie}} \mybox{color0}{\strut{.}}\\
\bottomrule
\end{tabular}
\caption{Saliency maps of BERT generated by three common gradient methods and by our proposed method, ALTI, for a \textbf{positive} sentiment prediction example of IMDB dataset.} 
\label{tab:long_imdb}
\end{table*}

\begin{figure*}[!h]
\centering
\begin{minipage}[b]{\textwidth}
\includegraphics[width=1\textwidth]{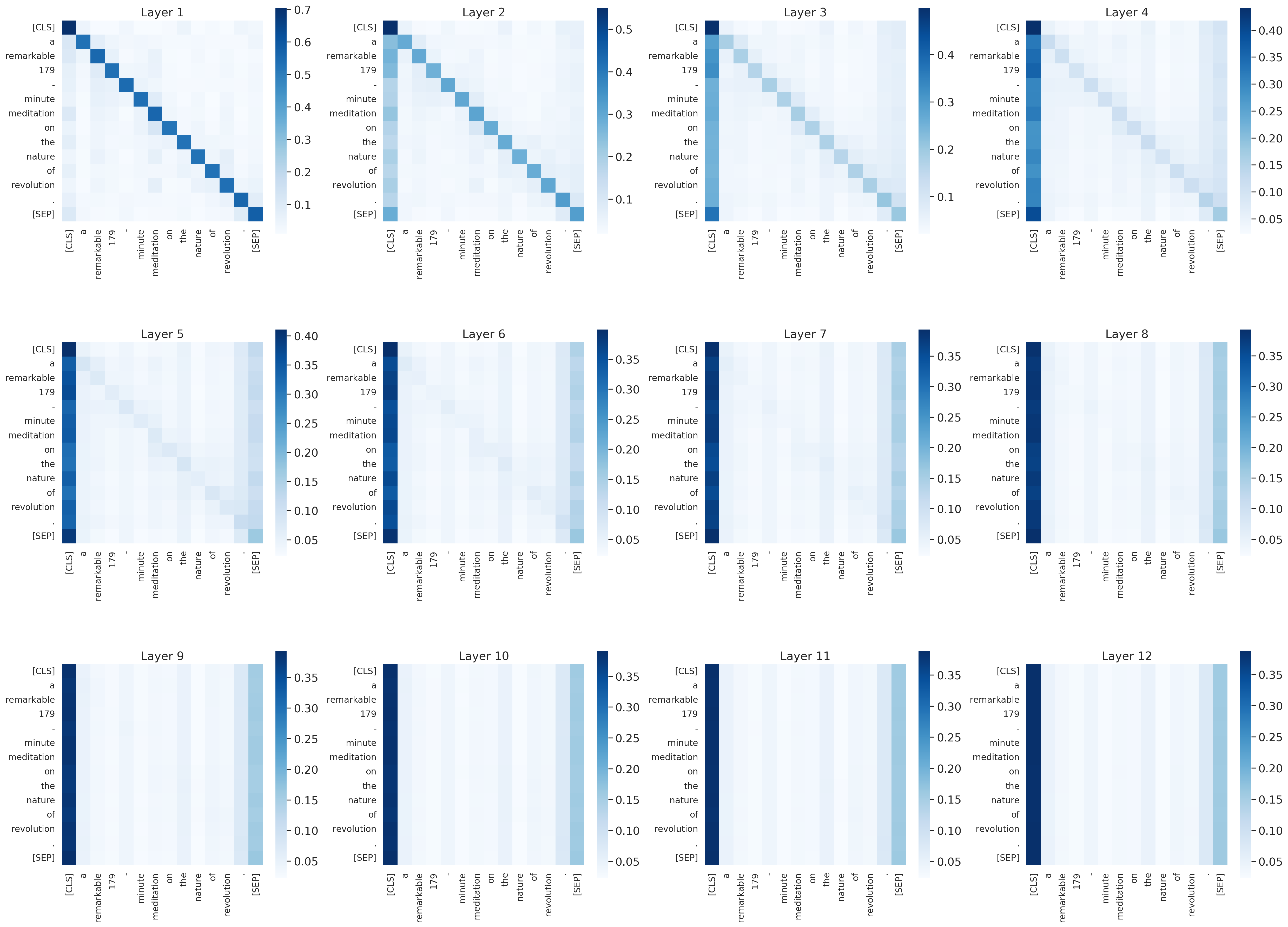}
\end{minipage}\hfill
\caption{Attn Rollout relevancies $\mathbf{R}^l$ in BERT across layers.}
\label{fig:sst2_bert_17_rollout}
\end{figure*}

\begin{figure*}[!h]
\centering
\begin{minipage}[b]{\textwidth}
\includegraphics[width=1\textwidth]{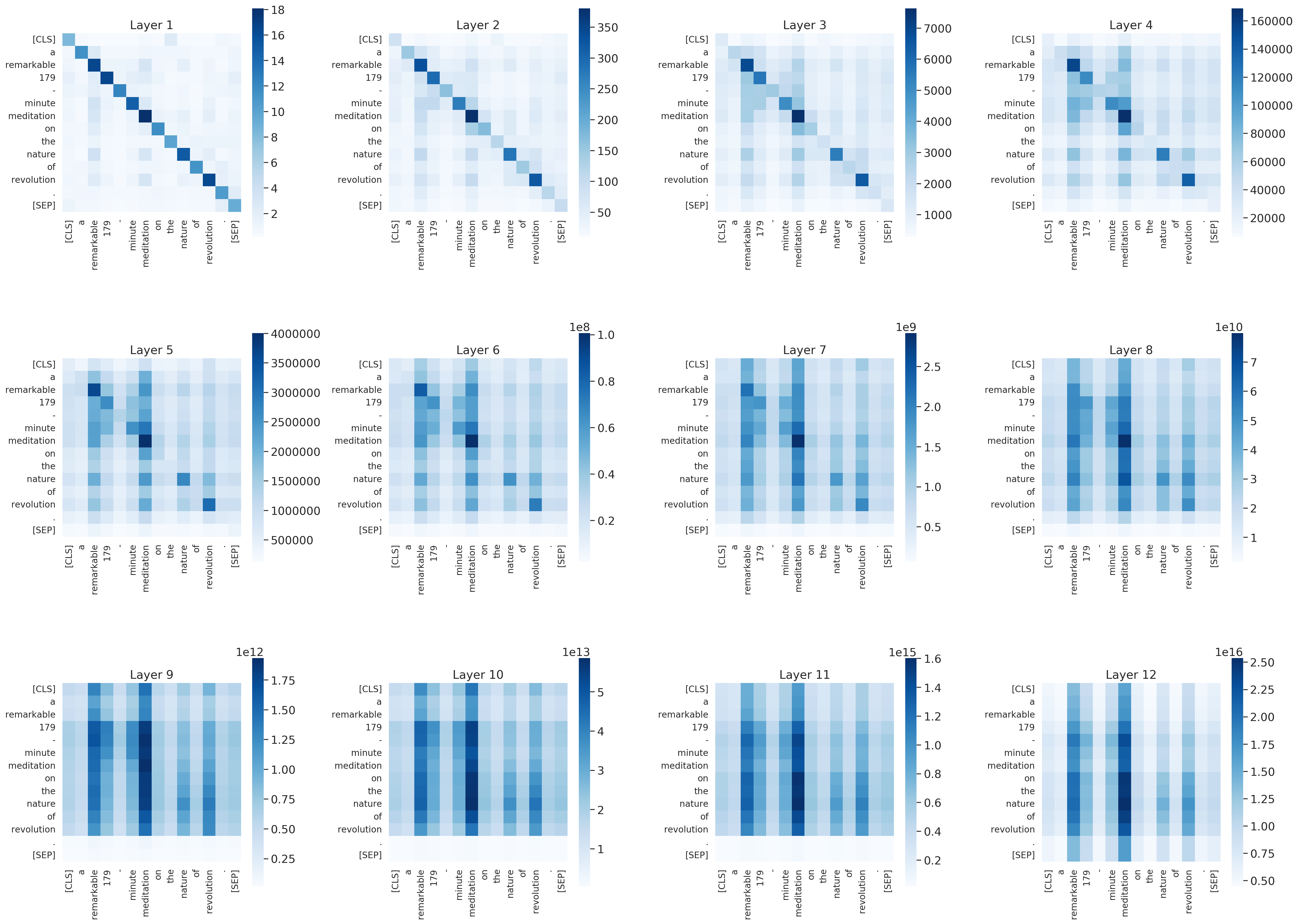}
\end{minipage}\hfill
\caption{Globenc relevancies $\mathbf{R}^l$ in BERT across layers.}
\label{fig:sst2_bert_17_norms}
\end{figure*}

\begin{figure*}[!h]
\centering
\begin{minipage}[b]{\textwidth}
\includegraphics[width=1\textwidth]{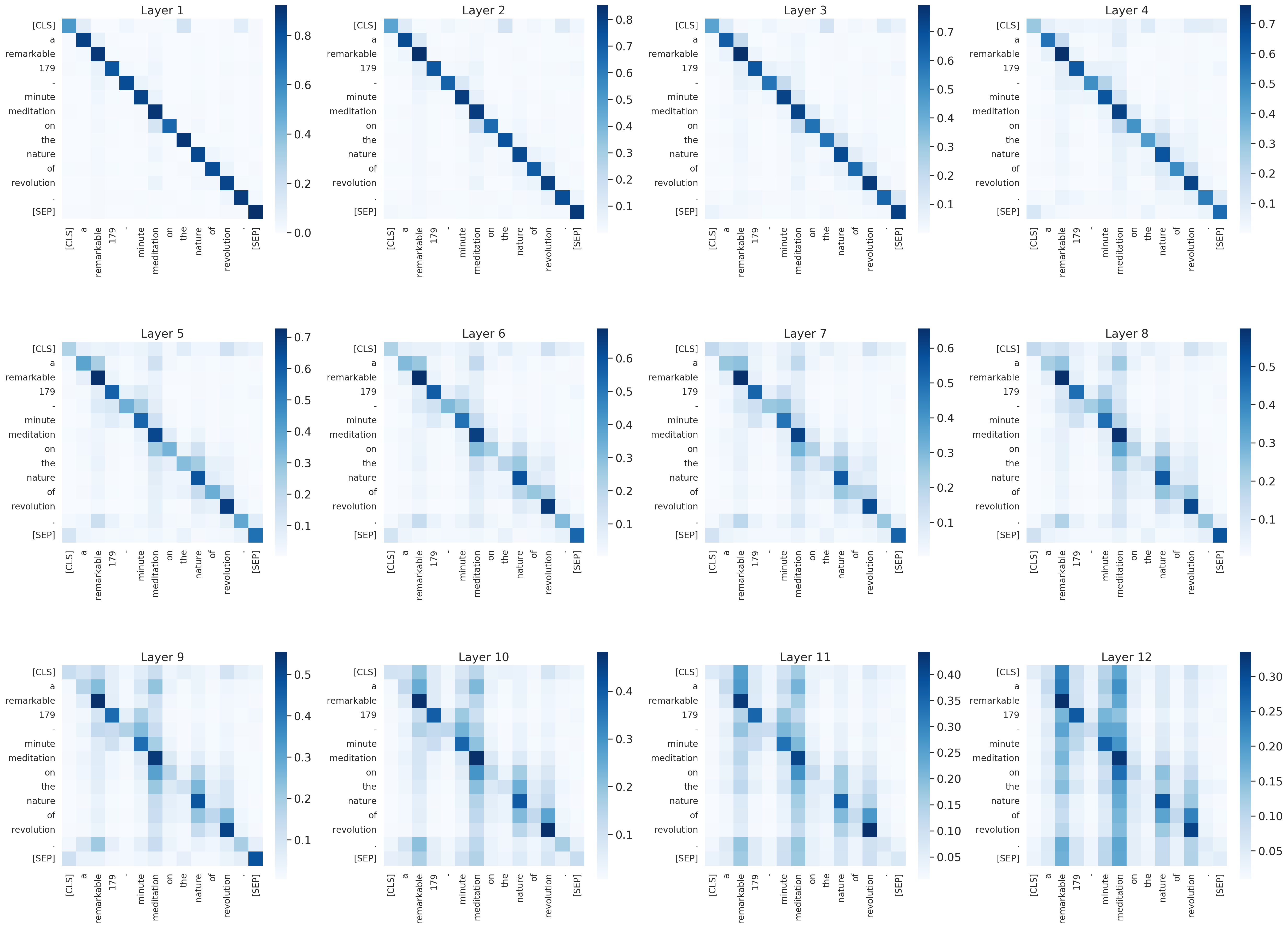}
\end{minipage}\hfill
\caption{ALTI method relevancies $\mathbf{R}^l$ in BERT across layers.}
\label{fig:sst2_bert_17_our}
\end{figure*}


\end{document}